\pdfoutput=1
\documentclass[11pt,usenames,dvipsnames]{article}
\usepackage{emnlp2022}
\usepackage{times}
\usepackage{latexsym}
\usepackage{tabularx}
\usepackage{multirow}
\usepackage{booktabs} 
\usepackage{flushend} 
\usepackage{balance} 
\usepackage{inconsolata}
\usepackage{graphicx}
\usepackage[utf8]{inputenc} 
\usepackage[T1]{fontenc}    
\usepackage{hyperref}       
\usepackage{url}            
\usepackage{booktabs}       
\usepackage{amsfonts}       
\usepackage{nicefrac}       
\usepackage{microtype}      
\usepackage{xcolor}         
\usepackage{multicol}
\usepackage{transparent}
\usepackage{array}
\usepackage{bm}
\usepackage{perpage} 
\MakePerPage{footnote} 
\usepackage{tabularx}
\usepackage{floatrow, makecell}
\usepackage{wrapfig,lipsum,booktabs}
\usepackage{pifont}
\usepackage{amssymb}
\usepackage{amsmath}
\usepackage{paralist}
\usepackage{caption}
\usepackage{subcaption}

\definecolor{midnightgreen}{rgb}{0.0, 0.29, 0.33}
\definecolor{lightgreen}{rgb}{0.0, 0.4, 0.20}
\definecolor{darkpink}{rgb}{0.91, 0.33, 0.5}
\definecolor{darkmagenta}{RGB}{139, 0, 139}

\newcommand{\model}[1]{ANCE-Tele}

\newcommand{\redbf}[1]{\textcolor{red}{\bf {#1}}}
\newcommand{\bluebf}[1]{\textcolor{blue}{\bf {#1}}}

\title{Reduce Catastrophic Forgetting of Dense Retrieval Training with Teleportation Negatives}

\author{
\textbf{Si Sun}$^{1}$, \textbf{Chenyan Xiong}$^{2}$, \textbf{Yue Yu}$^3$, \textbf{Arnold Overwijk}$^2$, \textbf{Zhiyuan Liu}$^{4,5}$, \textbf{Jie Bao}$^{1}$ \\
$^1$Dept. of Electron. Eng., Tsinghua University, Beijing, China\\
$^2$Microsoft Research, Redmond, USA \quad $^3$Georgia Institute of Technology, Atlanta, USA \\
$^4$Dept. of Comp. Sci. \& Tech., Institute for AI, Tsinghua University, Beijing, China \\
$^5$Beijing National Research Center for Information Science and Technology \\
\texttt{s-sun17@mails.tsinghua.edu.cn}; \texttt{\{chenyan.xiong, arnold.overwijk\}@microsoft.com} \\
\texttt{yueyu@gatech.edu}; \texttt{\{liuzy, bao\}@tsinghua.edu.cn}
}

\begin{document}
\maketitle
\begin{abstract}
In this paper, we investigate the instability in the standard dense retrieval training, which iterates between model training and hard negative selection using the being-trained model.
We show the catastrophic forgetting phenomena behind the training instability, where models learn and forget different negative groups during training iterations. We then propose ANCE-Tele, which accumulates momentum negatives from past iterations and approximates future iterations using lookahead negatives, as ``teleportations'' along the time axis to smooth the learning process. On web search and OpenQA, ANCE-Tele outperforms previous state-of-the-art systems of similar size, eliminates the dependency on sparse retrieval negatives, and is competitive among systems using significantly more (50x) parameters.
Our analysis demonstrates that teleportation negatives reduce catastrophic forgetting and improve convergence speed for dense retrieval training. Our code is available at~\url{https://github.com/OpenMatch/ANCE-Tele}.
\end{abstract}
\section{Introduction}

Dense retrieval (DR) learns to represent data into a continuous representation space and matches user information needs (``query'') with target information (``document'') via efficient nearest neighbor search~\citep{huang2013learning, lee2019latent}. 
Recent research shows strong empirical advantages of dense retrieval in various information access scenarios, such as OpenQA~\citep{dpr}, web search~\citep{ance},  and conversational search~\citep{yu2021convdr}.

A unique challenge of dense retrieval is in the selection of training negatives~\citep{dpr}. 
For each query, dense retrieval models need to distinguish a few relevant documents from all other negative documents in the entire corpus, the latter often beyond the scale of millions, making negative sampling a necessity. At the same time, the nature of retrieval makes random negatives trivial and uninformative~\citep{ance}, making effective negative sampling difficult.
Recent research addressed this challenge using an iterative training approach: first training using negatives generated by sparse retrieval for a while, then refreshing the training negatives using the being-trained DR models themselves, and alternating the two phases till convergence~\citep{ance}.

The iterative training-and-negative-refreshing approach yields strong results and became a standard in many DR systems~\citep[e.g.]{tas-b, ren2021rocketqav2, cocondenser}.
However, the refresh of hard negatives may change the learning landscape too dramatically and cause optimization issues. Many found the little benefit of training more than one refresh~\citep{ouguz2021domain, cocondenser} and the significant fluctuations in the model accuracy with more iterations due to training instability~\citep{ance}.


\begin{figure*}[t]
\centering
\includegraphics[width=\textwidth]{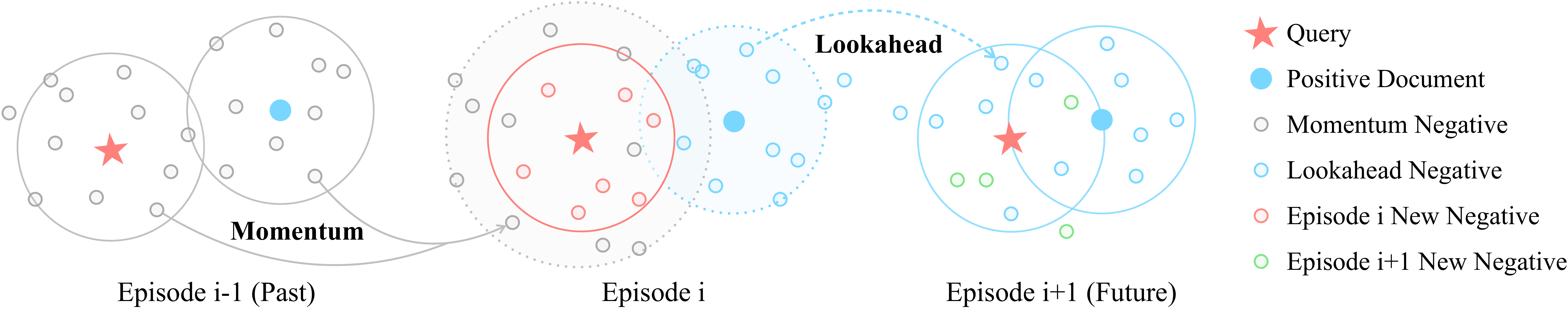}
\caption{\label{fig:framework} Illustration of mining teleportation (momentum and lookahead) negatives in episode $i$ of \model{}.}
\vspace{-0.2cm}
\end{figure*}
\vspace{-0.02cm}


In this paper, we aim to address the instability issue in dense retrieval training.
We first conduct a deep investigation on the model behaviors during training and reveal a phenomenon behind the instability: catastrophic forgetting~\citep{kirkpatrick2017overcoming}.
After training with refreshed negatives in one episode, the dense retriever's accuracy drops on a large fraction of \textit{training queries} (20-30\%) that have been learned earlier.
Our analysis shows that negatives refreshed in different iterations may differ drastically, e.g., covering distinct distractions, and model training ``swings'' between these distractor groups, rather than capturing them all together.


With these observations, we develop \model{}, which upgrades ANCE~\citep{ance}, a standard dense retrieval training strategy, using \textit{teleportation negatives} with momentum and lookahead mechanisms. Specifically, momentum negatives record the negatives encountered in past iterations, while lookahead negatives use the neighbors of positive documents as surrogates to approximate the negatives appearing in the future. These teleportation negatives smooth out the training process of dense retrieval. Figure~\ref{fig:framework} illustrates the training process of \model{}.

In our experiments on web search and OpenQA, \model{} outperforms previous state-of-the-art systems of similar parameter sizes,  without combining negatives from sparse retrieval or distillation from cross-encoder teachers.
Simply including the teleportation negatives in training enables a BERT-base sized dense retriever (110 million parameters) to outperform the recent state-of-the-art model with five billion parameters.

Our analysis on the learning dynamics of \model{} confirmed the benefits of teleportation negatives.
The momentum negatives from previous iterations reduce the catastrophic forgetting issue and improve learning stability, as they note the model to ``remember'' previously learned signals. 
The lookahead negatives sampled from the neighborhood of the positive documents provide an efficient forecast of future hard negatives and also  improve the convergence speed.
The two together improve the stability, efficiency, and effectiveness of dense retrieval training, while imposing zero GPU computation overhead.

After recapping related work, we investigate the instability issue of the iterative training process in Section~\ref{sec:stable-analyses} and present \model{} in Section~\ref{sec:tele-neg-sample}. 
Experimental settings and evaluations are discussed in Section~\ref{sec:exp-method} and Section~\ref{sec:eval-res}.
We conclude in Section~\ref{sec:conclusion}.

\section{Related Work}

The classic idea of learning representations for retrieval~\citep[e.g.]{deerwester1990indexing, huang2013learning} is recently revived with pretrained language models~\citep{devlin2019bert}.
\citet{lee2019latent} first presented the BERT-based dual-encoder dense retrieval formulation.
\citet{dpr} introduced BM25 negatives in DR training.
\citet{ance} showed the necessity of hard negatives and introduced the iterative training-and-negative-refreshing process.
These techniques formed the basic dense retrieval setup that achieves strong performances on a wide range of scenarios.

To improve dense retrieval training strategy is an active research front. \citet{zhan2021optimizing} mitigated the risk of the delayed negative refresh by mining ``real-time'' hard negatives with a step-wise updating query encoder and a fixed document index. \citet{hofstatter2021efficiently} observed that the iterations may lead to a fragile local optimal and balanced the negatives among query clusters.
\citet{rocketqa} filtered the hard negatives using a stronger cross-encoder ranking model.
With access to query-document term interactions, the ranking models are more powerful and can enhance DR training via knowledge distillation as well~\citep{lin2020distilling, hofstatter2021efficiently, ren2021rocketqav2}.
\citet{lewis2021boosted} trained a series of dense retrievers through boosting and improved retrieval accuracy under approximate nearest neighbor search (ANN).

Recent research also identified several mismatches between pretrained models and DR.
One mismatch is the locality of token level pretraining versus the needs of full sequence embeddings in DR~\citep{lu2021less}, which can be reduced by enforcing an information bottleneck on the sequence embedding in pretraining~\citep{wang2021tsdae, gao-callan-2021-condenser}.
Another is the lack of alignment in the pretrained sequence representations, which can be improved using sequence contrastive pretraining~\citep{meng2021coco,cocondenser}. Pretrained models with billions or more parameters also improve DR accuracy, though the benefit of scaling in DR is less than observed in other tasks~\citep{gtr, neelakantan2022text}.

DR often serves as the first stage retrieval in many language systems. Jointly training DR with later stage models can lead to better accuracy, with labels and signals from more sophisticated models in later stages, for example, question answering systems~\citep{izacard2020distilling, zhao2021distantly} and reranking models~\citep{zhang2021adversarial}.

\section{Dense Retrieval Training Analysis}
\label{sec:stable-analyses}

In this section, we first present the preliminaries of a standard dense retrieval setup and then investigate its training instability issues.

\subsection{Preliminaries on Dense Retrieval}

The first stage retrieval task is to find a set of relevant documents $D^+$ from a corpus $C$, for a given query $q$. The efficiency constraints often require the retrieval system to first represent query and documents {independently} into a vector space and then match them by efficient vector similarity metrics.
Dense retrieval refers to methods that use a continuous dense vector space, in contrast to the discrete bag-of-word space used in sparse retrieval.

\textbf{Bi-Encoder Model.}
A standard formulation of DR is to embed query and documents using dual/bi-encoders initialized from pretrained language models~\citep{lee2019latent}:
\begin{align}
    f(q, d; \theta) &= g(q; \theta) \cdot g(d; \theta). \label{eq.dualencoder}
\end{align}
The encoder $g(\circ,\theta)$ embeds $q$ and $d$ using its parameter $\theta$. The match uses dot product ($\cdot$). 

Retrieval in the embedding space with common similarity metrics, such as cosine, dot product, L2 distance, is supported by fast approximate nearest neighbor (ANN) search~\citep{sptag,scann,faiss}. 
For example, we can retrieve the top K documents for a given $q$ from the ANN index with high efficiency at a small cost of exactness:
\begin{align}
    \text{ANN}_{f(q,\circ; \theta)} &= \text{Top K}^\text{ANN}_{d \in C} f(q, d; \theta).
\end{align}

\textbf{Iterative Training.} The training labels for retrieval are often provided as a set of relevant documents $D^+$ for each $q$, clicked web documents, passages containing the answer, etc.
As the retrieval model needs to separate relevant documents $d^+$ from the entire corpus, all the rest corpus $C\setminus D^+$ are negatives, which are often too many to enumerate and require sampling. 

Another nature of retrieval is that most irrelevant documents are trivial to distinguish. Only a few are challenging. It is unlikely for random sampling to hit these hard ones and produce informative training negatives.
A widely used approach~\citep[e.g.,]{ance, ren2021rocketqav2, oguz2020unik} is to first train with hard negatives from sparse retrieval's top results, then use the trained DR model to refresh the hard negatives (self negative), and alternate through this training-and-negative-refreshing circle.

Without loss of generality, for a given training query $q$ and its relevance documents $D^+$, one training iteration $i$ in this training process is to find $\theta^*_i$ that minimizes the following loss:
\begin{align}
    &\mathcal{L}_i = \sum_{q; d^+ \in D^+} \sum_{d^- \sim D^-_{i}}l(f(q, d^+; \theta_i), f(q, d^-; \theta_i)); \nonumber \\
    &D_i^- = \begin{cases}
    \text{ANN}_{f(q,\circ; \theta^*_{i-1})} \setminus D^+, & i>1\\
    \text{BM25}(q,\circ)  \setminus D^+ & i=1. \label{eqn:ance}
    \end{cases}
\end{align}

The negatives $D_i^-$ are constructed using model checkpoint from past episode or warmed up using sparse retrieval (BM25)~\citep{ance}, where $d^-$ are sampled uniformly without replacement. 
The model is trained to separate $D_i^-$ from $d^+$ using the ranking loss $l$,   e.g., cross entropy or hinge loss.
For clarity, we refer to an iteration of negative construction and model training as one \textit{training episode} in the rest of this paper, with episode $1$ referring to the first fine-tuning iteration from the pretrained model.

\subsection{Learning Instability Investigation}
\label{sec:learn-instable}

We now investigate the training process described in Eqn.~\ref{eqn:ance}. We first study the training curves of various iterative training configurations, proceed with model behaviors, and then the dynamics of hard negative selection through training episodes.

\textbf{Analysis Setting.} We use MS MARCO passage retrieval benchmark~\citep{bajaj2016ms} and the iterative training configurations from ANCE, starting from BM25 negatives and then self-mined negatives~\citep{ance}. 
In addition, we conduct ANCE training from both vanilla BERT~\citep{devlin2019bert} and coCondenser~\citep{cocondenser}. The latter continues pretraining with sequence contrastive learning for dense retrieval.

The iterative training can be viewed as a continual learning process~\citep{mitchell2018never}, with each negative refresh forms a mini-task with new training signals.
To better account for the training negative changes, we also experiment with cyclical learning rate scheduling~\citep{smith2017cyclical}, which  warms up for each episode individually. We tune the hyperparameters to make the training as stable as possible. The detailed configurations in our analysis are listed in Appendix~\ref{app:ance-ana}. 

\begin{figure}[t]
\centering 
    \begin{subfigure}[t]{0.48\linewidth}
        \includegraphics[height=3.0cm]{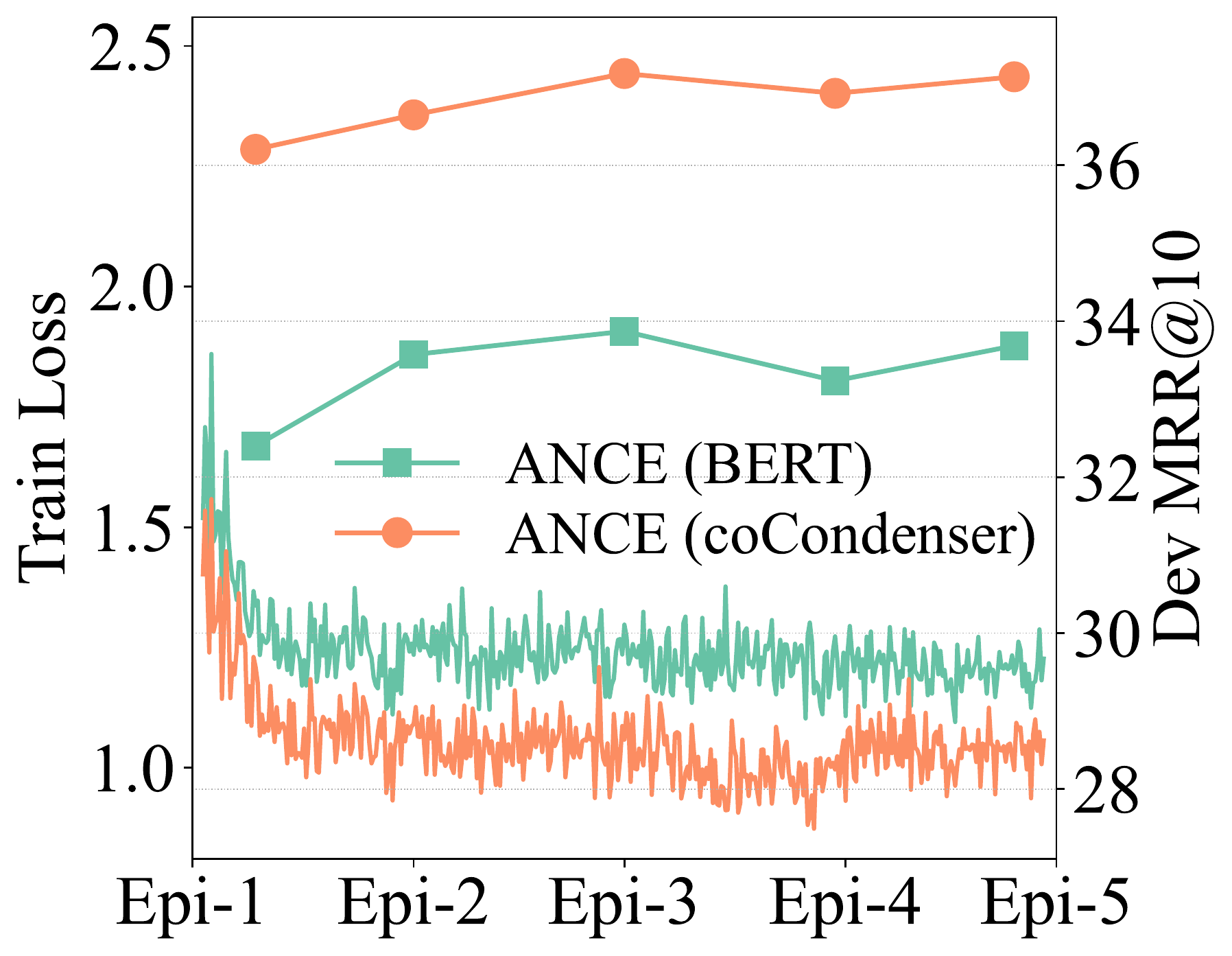}
        \caption{Standard ANCE. \label{fig:vanilla-ance}}
    \end{subfigure}~
    \begin{subfigure}[t]{0.48\linewidth}
        \includegraphics[height=3.0cm]{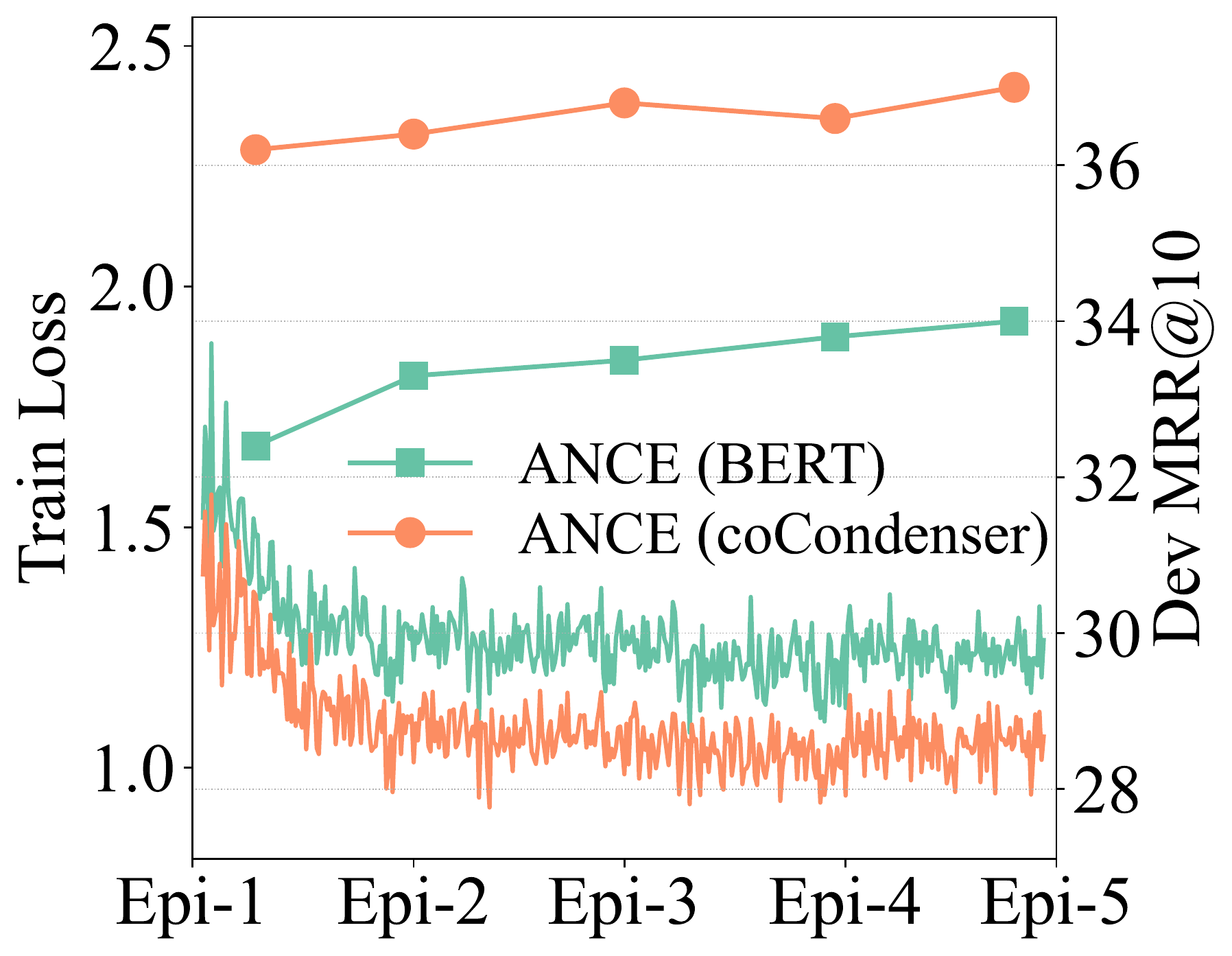}
        \centering
        \caption{ANCE with CyclicLR.\label{fig:re-warmup-ance}}
    \end{subfigure}
    \vspace{-0.2cm}
\caption{\label{fig:ance_perform} Training and testing curves in ANCE training episodes on MS MARCO. We use standard ANCE with linear decay in (a) and our tuned Cyclical LR in (b).} 
\end{figure}


\begin{figure}[t]
  \centering
  \vspace{-0.2cm}
    \begin{subfigure}[t]{0.48\linewidth}
        \includegraphics[height=4.0cm]{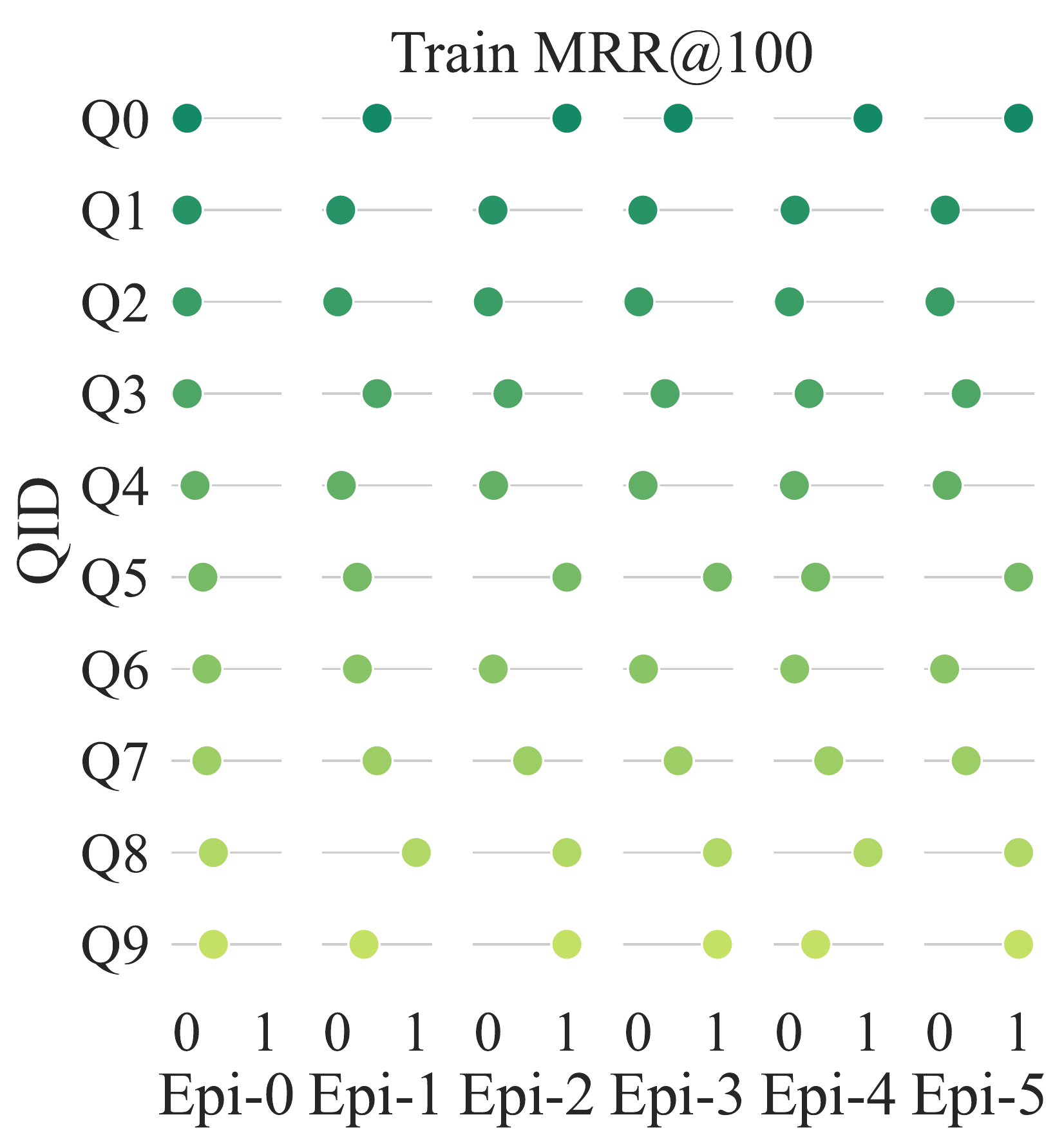}
        \caption{ANCE (BERT). \label{fig:forget-case-bert}}
    \end{subfigure}
    \begin{subfigure}[t]{0.48\linewidth}
        \includegraphics[height=4.0cm]{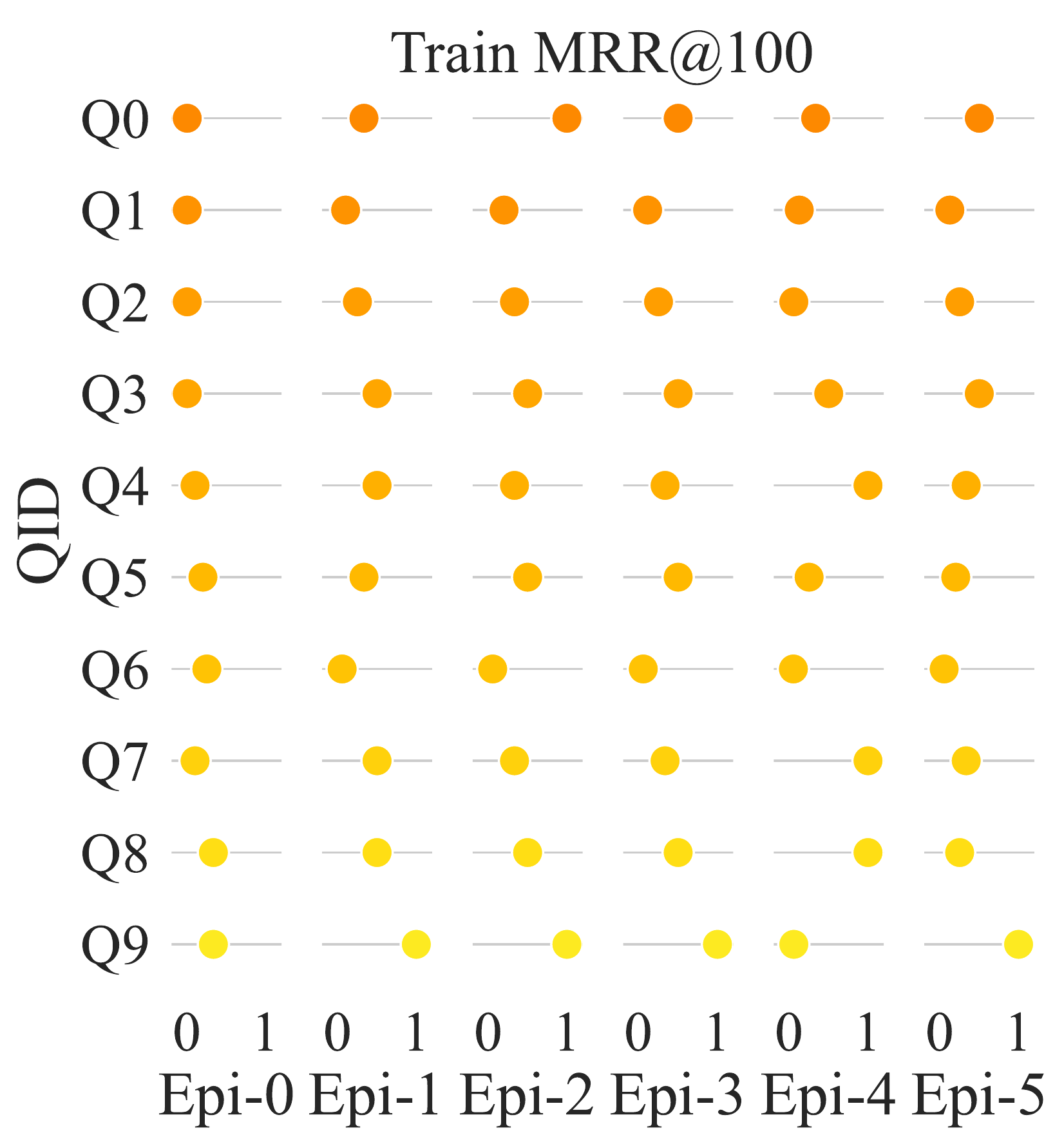}
        \centering
        \caption{ANCE (coCondenser). \label{fig:forget-case-coco}}
    \end{subfigure}
    \vspace{-0.2cm}
\caption{\label{fig:forget-example} Accuracy on ten random \textit{training} queries during ANCE training with CyclicLR. 
}
\end{figure}


\begin{figure}[t]
  \centering
    \begin{subfigure}[t]{0.48\linewidth}
        \includegraphics[height=3cm]{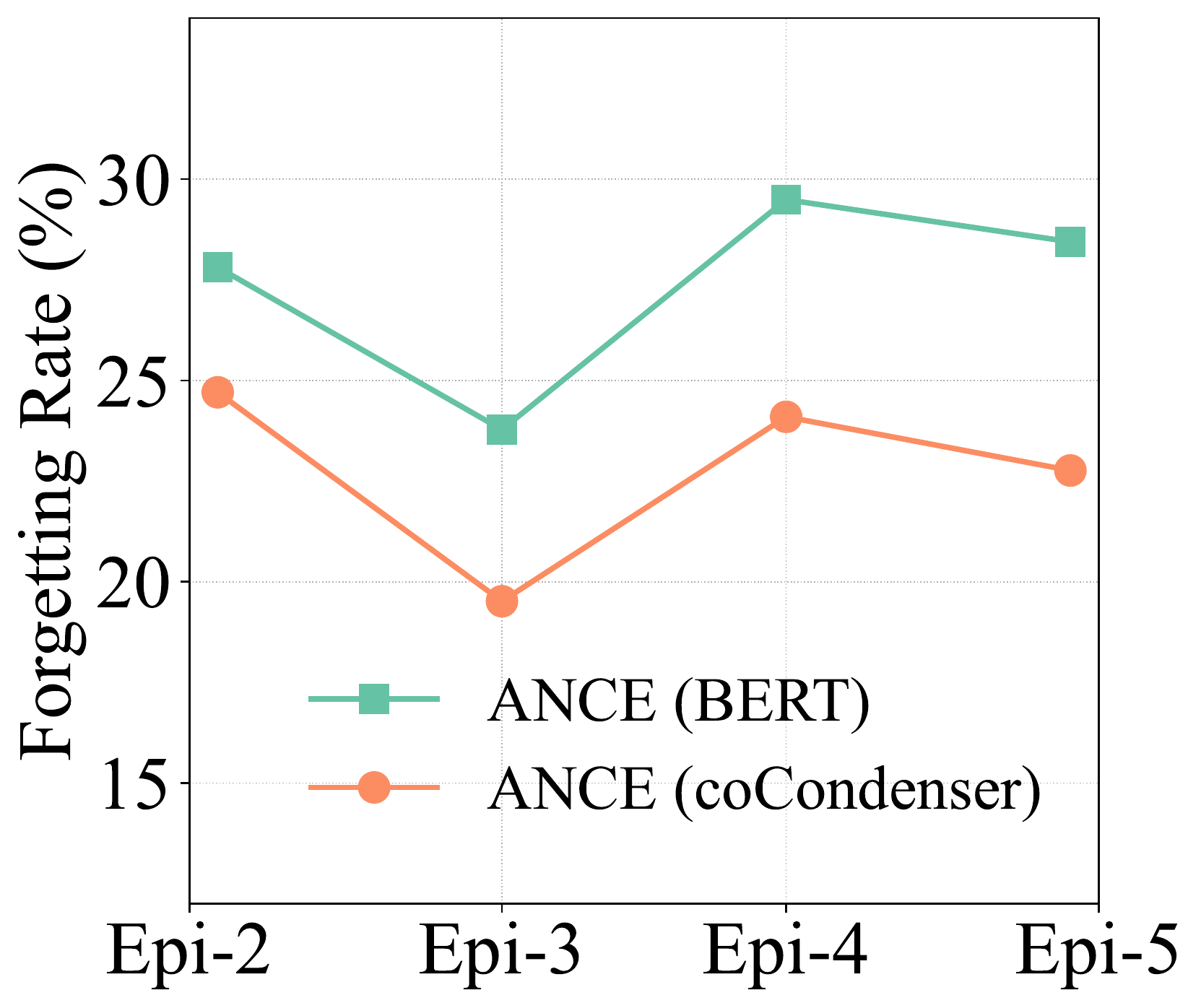}
        \caption{Standard ANCE.\label{fig:vanilla-ance-forget-rate}}
    \end{subfigure}
    \begin{subfigure}[t]{0.48\linewidth}
        \includegraphics[height=3cm]{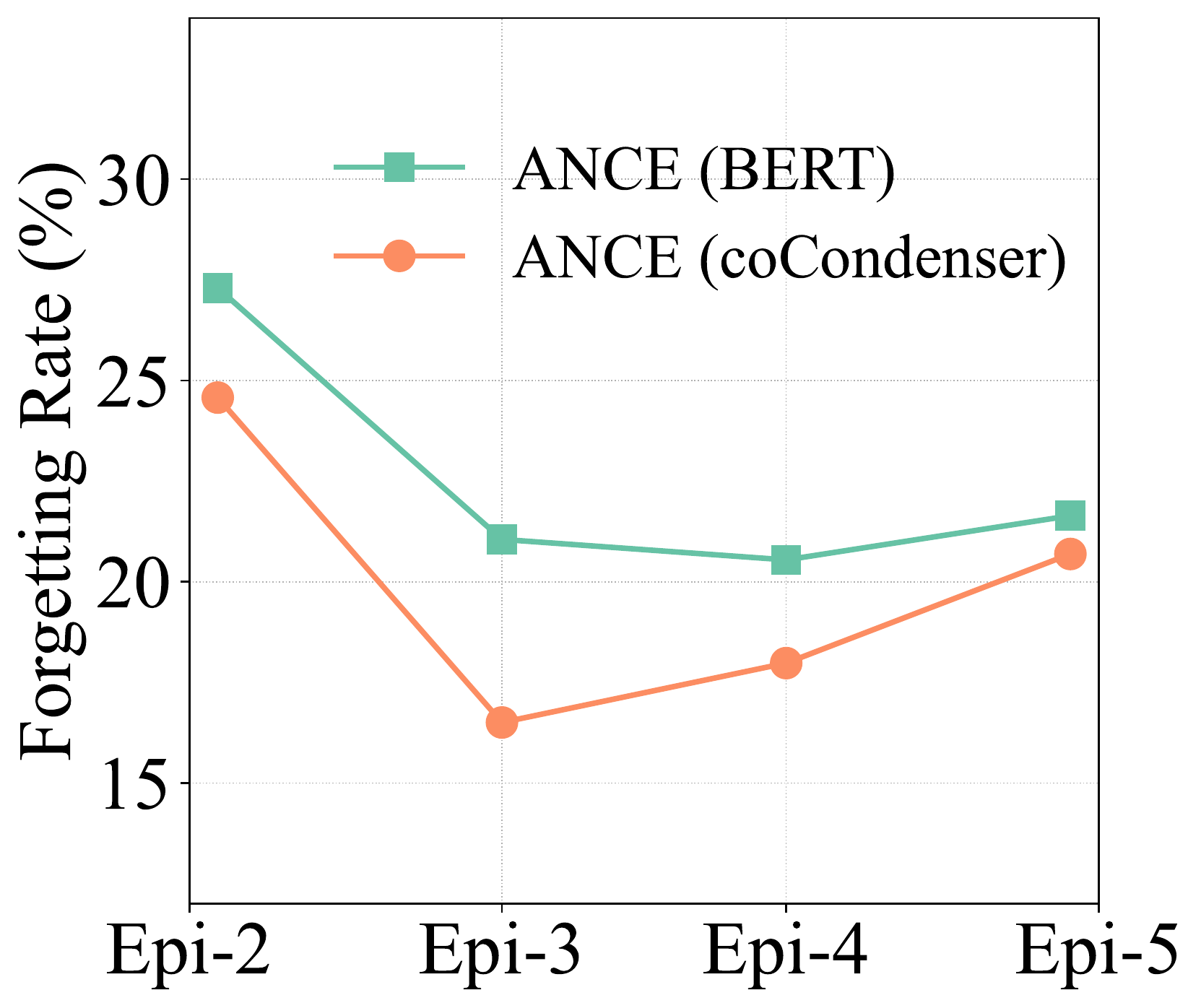}
        \centering
        \caption{ANCE with CyclicLR.\label{fig:warmup-ance-forget-rate}}
    \end{subfigure}
    \vspace{-0.2cm}
\caption{\label{fig:ance-forget-rate} Fraction of training queries the models performed worse after one training episode, compared with their MRR@100 in the past episode.}
\end{figure}

\begin{figure}[t]
  \centering
    \begin{subfigure}[t]{0.48\linewidth}
        \includegraphics[height=3.2cm]{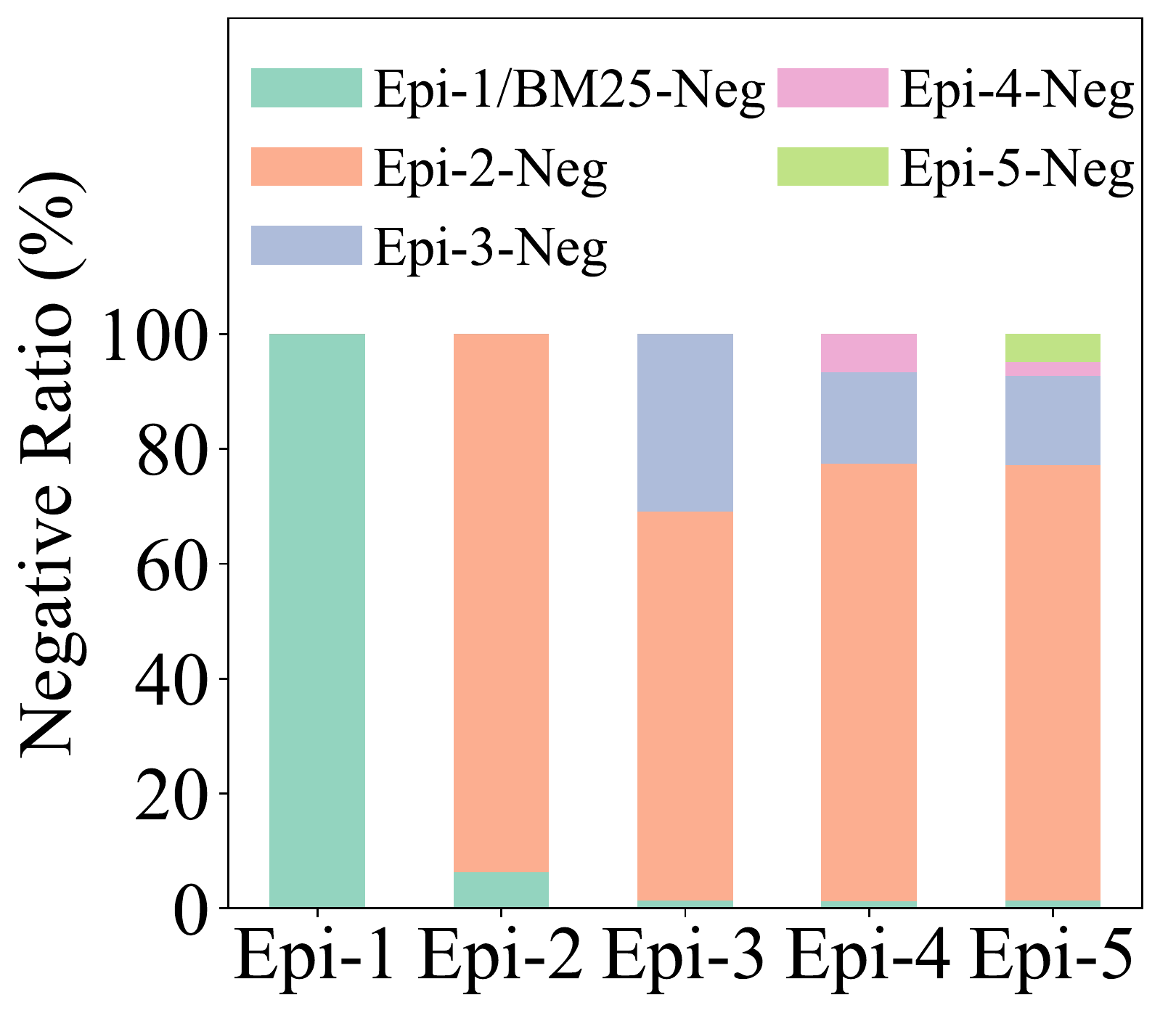}
        \caption{ANCE (BERT).}
    \end{subfigure}
    \begin{subfigure}[t]{0.48\linewidth}
        \includegraphics[height=3.2cm]{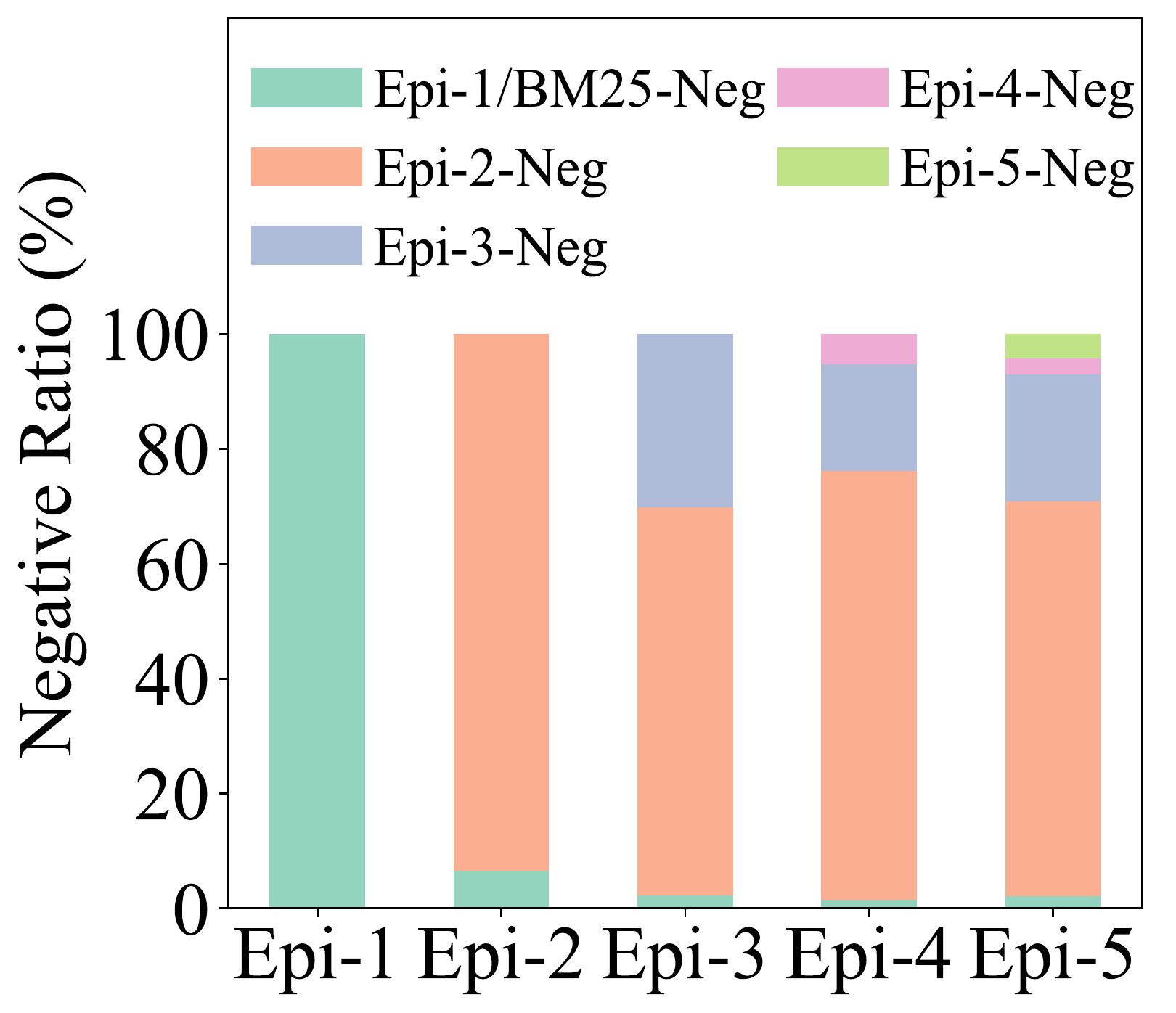}
        \centering
        \caption{ANCE (coCondenser).\label{fig:neg_frac}}
    \end{subfigure}
    \vspace{-0.2cm}
\caption{\label{fig:neg_compose} Composition of all negatives in each episode of ANCE (BERT/coCondenser) with CyclicLR.
}
\end{figure}



\begin{figure*}[t]
  \centering
      \begin{subfigure}[t]{0.24\linewidth}
        \includegraphics[height=3.0cm]{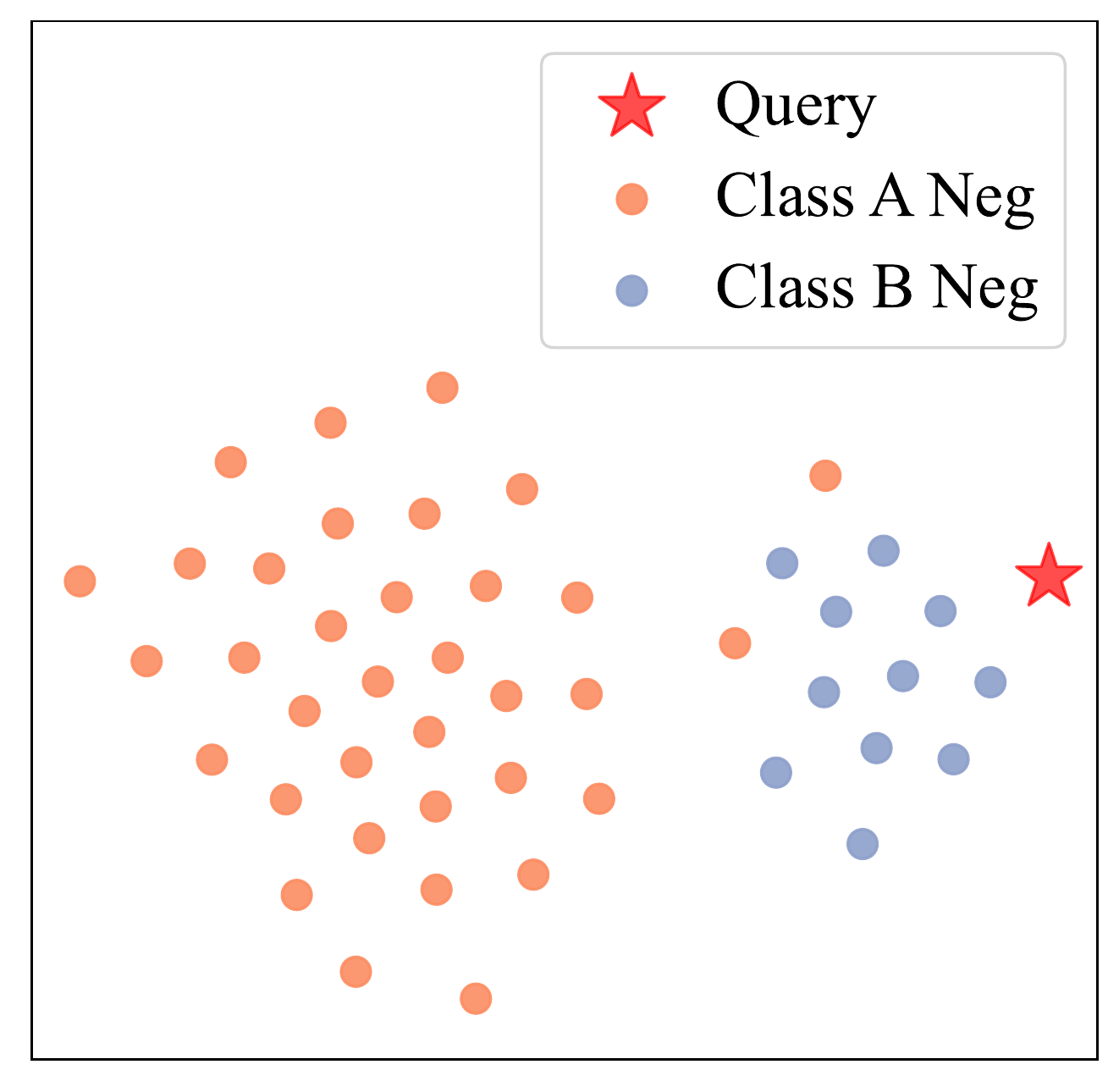}
        \caption{Epi-2.}
    \end{subfigure}
    \hspace{0.1cm}
    \begin{subfigure}[t]{0.24\textwidth}
        \includegraphics[height=3.0cm]{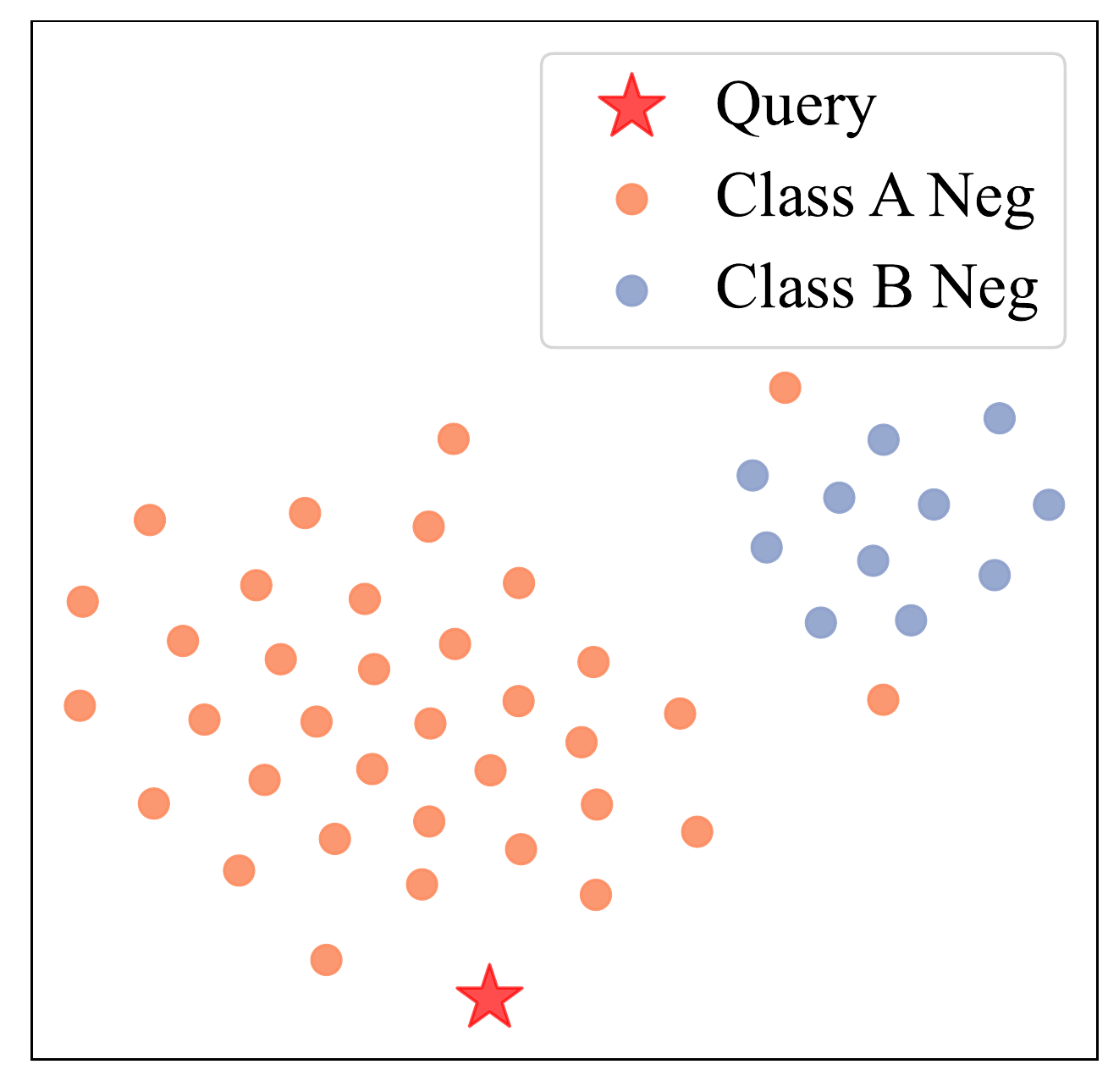}
        \caption{Epi-3.}
    \end{subfigure}
    \hspace{0.1cm}
    \begin{subfigure}[t]{0.24\linewidth}
        \includegraphics[height=3.0cm]{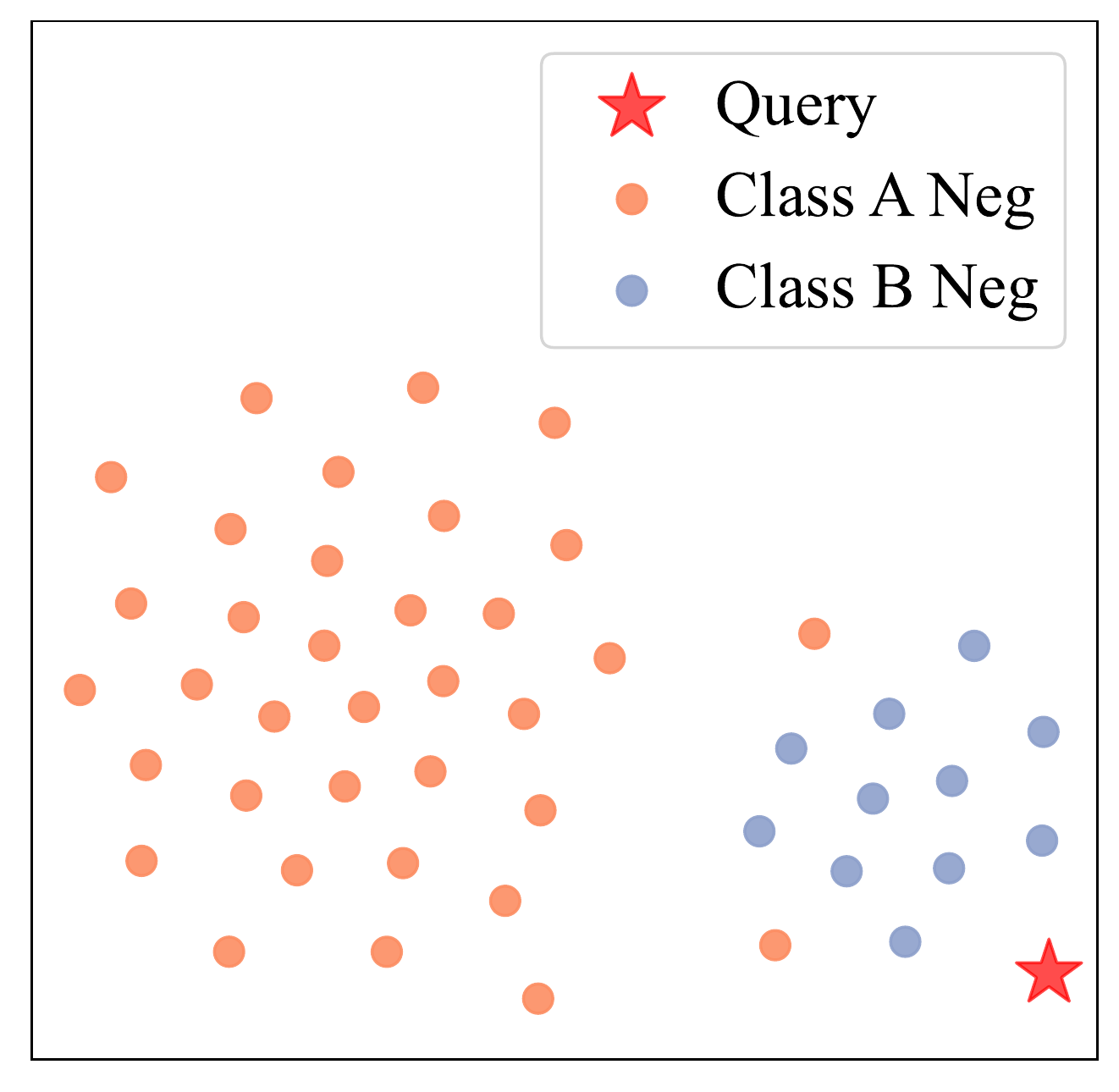}
        \caption{Epi-4.}
    \end{subfigure}
    \hspace{-0.3cm}
    \begin{subfigure}[t]{0.24\linewidth}
        \includegraphics[height=3.0cm]{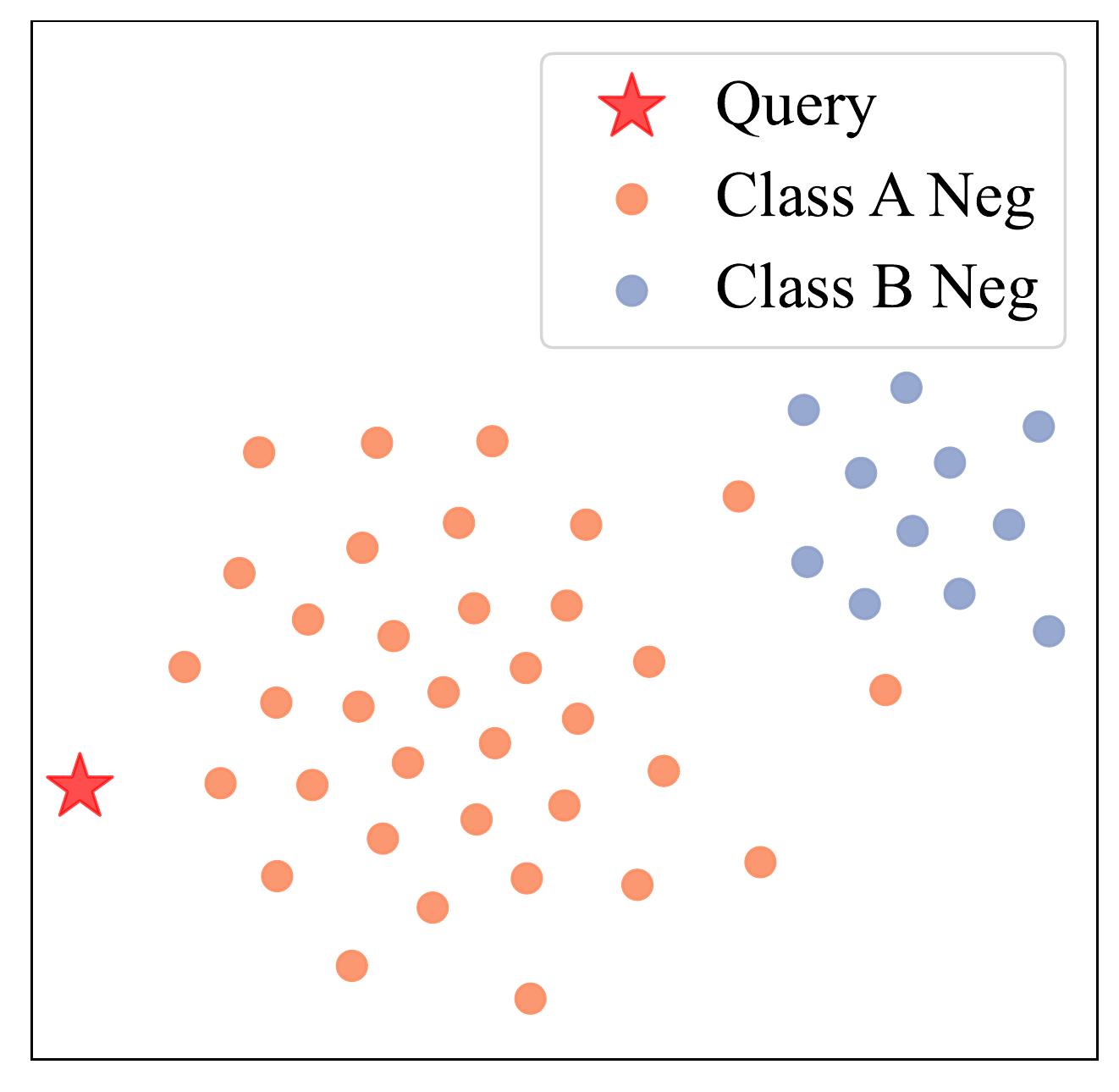}
        \centering
        \caption{Epi-5.}
    \end{subfigure}
    \hspace{-0.4cm}
    \vspace{-0.1cm}
\caption{The t-SNE plots for negative swing. Query: \textit{most popular breed of rabbit}. Its MRR@100 jumps up and down during iterative training: Epi-1 (0.14); Epi-2 (0.08 $\downarrow$); Epi-3 (0.13 $\uparrow$); Epi-4 (0.11 $\downarrow$); Epi-5 (0.13 $\uparrow$). The details of class A/B negatives and more swing cases are shown in Table~\ref{app-tab:ance-swing-cases} (appendix). \label{fig:ance_tsne}}
\end{figure*}


\textbf{Learning Curves.} The training losses and development set accuracy (MRR@10) of ANCE variations are plotted in Figure~\ref{fig:ance_perform}. In standard ANCE, the training losses often increase in the beginning of episodes. This is designed as ANCE is to find hardest negatives for the model. However, it is undesired that the retrieval accuracy on Dev also fluctuates across the episodes. Using retrieval-oriented pretraining model, ANCE (coCondenser) elevates overall accuracy, but does not eliminate instability. 

Our well-tuned cyclical learning rate makes the training smoother but at a notable cost. ANCE with CyclicLR only reached similar performance with standard ANCE after five episodes and its peak accuracy is slightly lower.
Despite our best effort, the performance of ANCE (coCondenser), still drops at Epi-4 before it climbs back at Epi-5.

\textbf{Catastrophic Forgetting.} To understand this instability, we randomly sampled ten training queries and tracked their performances in Figure~\ref{fig:forget-example}. 

The examples indicate that the fluctuation in DR training is not due to overfitting, but catastrophic forgetting~\citep{kirkpatrick2017overcoming}, where models forget the examples they have learned in previous training steps, a common challenge in continual learning. The per query MRR jumps up and down during training and peaks at variant episodes. This shows the training instability problem is more severe than the averaged accuracy suggested.



In Figure~\ref{fig:ance-forget-rate}, we plot the average forgetting ratio of ANCE variants on all MARCO training queries. With standard ANCE training, a model may forget a startling 20-30\% of \textit{training} queries after one hard negative refresh!
At the cost of slower convergence, CyclicLR mitigates forgetting rate to around 20\%, except at Epi-2, when ANCE switches from BM25 negatives to self-negatives.

\textbf{Dynamics of Hard Negatives.} We keep analyzing the forgetting issue by investigating the training negatives used in each ANCE episode---the biggest moving piece in iterative training.
Figure~\ref{fig:neg_compose} presents the composition of training negatives of two ANCE variants with CyclicLR. Results for more variants can be found in Appendix~\ref{app:ance-ana}.



The dynamics of negatives in Figure~\ref{fig:neg_compose} reveal two behaviors of dense retrieval training: BM25 Warm Up Only and Negative Swing.

\textit{BM25 Warm Up Only.} 
After the warm-up stage (Epi-1), BM25 Negatives are discarded quickly and at near entirety. This echos previous observation that dense retrieval disagrees significantly with sparse retrieval~\citep{ance}.
Negatives informative for one side are trivial for the other.

\textit{Negative Swing.} The models swing between the negatives introduced at different episodes, rather than capturing all of them together. The self negatives introduced in episode $i$ often reduced at the next episode $i$+1. The model has learned in episode $i$ and some of negatives are no longer hard negatives at episode $i$+1. However, in episode $i$+2, some already captured episode $i$ negatives reappear as hard negatives, showing that the model forgets the information captured in episode $i$ when learning in episode $i$+1 negatives. The model seems to swing back and forth between several learning mods and reflects the catastrophic forgetting behavior.


To further illustrate the negative swing, we visualize an example query and its negatives during ANCE learning via t-SNE~\citep{tsne} in Figure~\ref{fig:ance_tsne}.
In this example there are two negative classes and the query is pushed between the two groups throughout the learning episodes. Capturing the information in one mod resulted in the forgetting of the other, as the model only sees one of them as hard negatives in each episode of ANCE training. 
Our further manual examination finds that for many queries with this negative swing behavior, the negative groups correspond to different common mistakes the retrieval system would make, for example, irrelevant documents that only cover part of the query. In Appendix~\ref{app:case} we show some example queries and negative groups of this behavior.

\section{Teleportation Negative Sampling}
\label{sec:tele-neg-sample}

In this section, we present \model{}, which introduces \textit{teleportation negatives} to ANCE training. Motivated by our analysis, \model{} unions the training negatives along the time-axis of ANCE episodes, with the goal of smoothing training signal changes, improving negative group coverage, and, ultimately, reducing catastrophic forgetting.

The construction of \textit{teleportation negatives} of \model{} is illustrated in Figure~\ref{fig:framework}. Specifically, the training episode $i$ is defined as:
\begin{align*}
    \mathcal{L}^\text{Tele}_i &= \sum_{q; d^+ \in D^+} \sum_{d^- \sim D^\text{Tele-neg}_{i}}l(f(q, d^+), f(q, d^-)).
\end{align*}
\model{} utilizes the same iterative training, but introduces two new components to standard ANCE negatives around $q$ (Eqn.~\ref{eqn.ance-neg}), momentum negatives (Eqn.~\ref{eqn.past-neg}) and lookahead negatives (Eqn.~\ref{eqn.future-neg}):
\begin{align}
    D_i^\text{Tele-neg} =   & \text{ANN}_{f(q,\circ; \theta^*_{i-1})}  \label{eqn.ance-neg} \\
     + & \alpha D_{i-1}^\text{Tele-neg}    \label{eqn.past-neg}     \\
    + & \beta \text{ANN}_{f(d^+,\circ; \theta^*_{i-1})}  \label{eqn.future-neg}
\end{align}

\textbf{Momentum negatives} include training negatives from past episodes using a momentum queue (Eqn.~\ref{eqn.past-neg}). They smooth out the training signal changes between past and current episodes.
A standard technique to reduce catastrophic forgetting~\citep{kirkpatrick2017overcoming}, including training signals from previous learning stages reminds the model to keep their learned knowledge.

\begin{figure}[t]
  \centering
   \vspace{-0.2cm}
    \hspace{-0.3cm}
    \begin{subfigure}[t]{0.48\linewidth}
        \includegraphics[height=3.2cm]{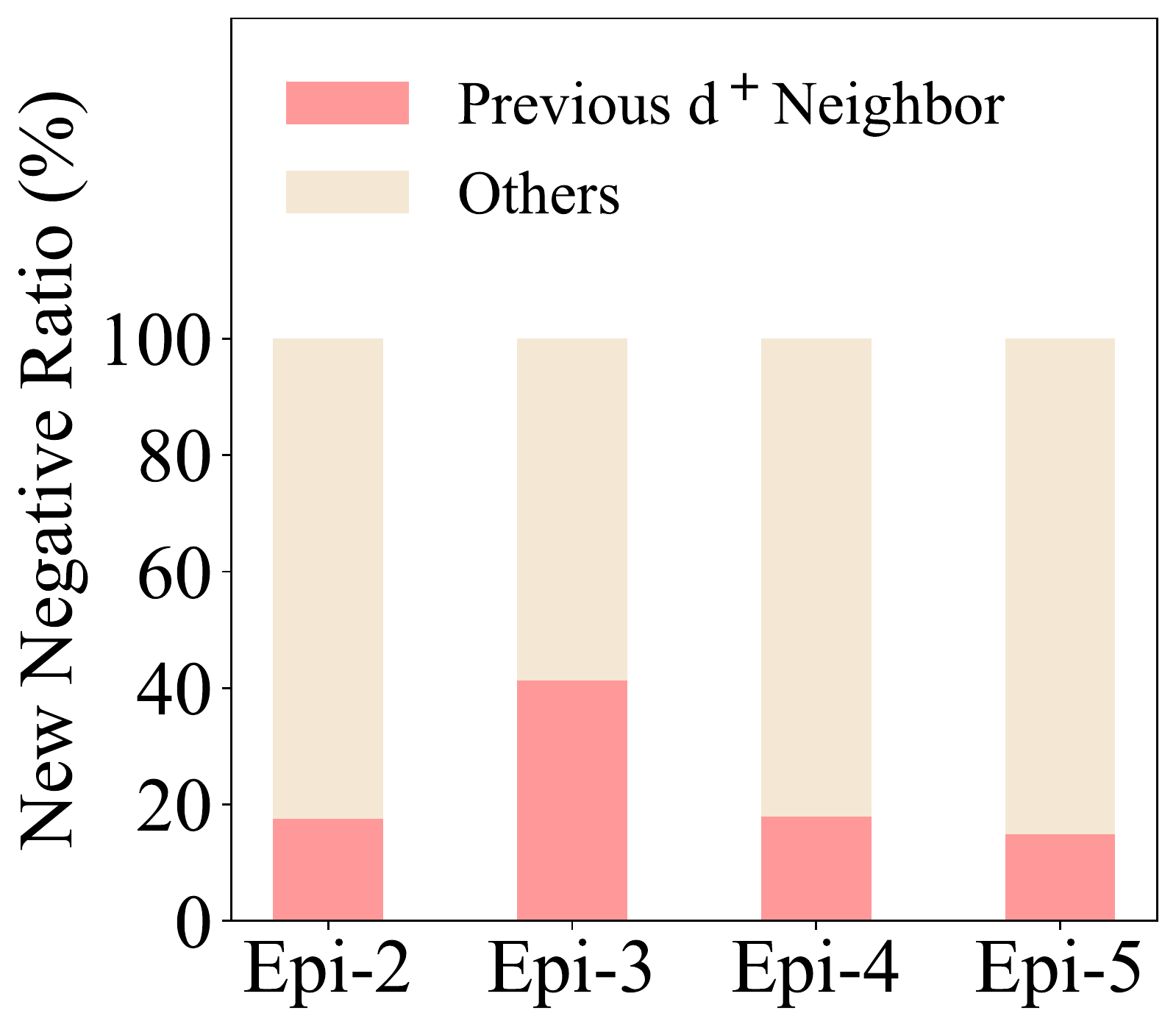}
        \caption{ANCE (BERT)}
    \end{subfigure}
    \begin{subfigure}[t]{0.48\linewidth}
        \includegraphics[height=3.2cm]{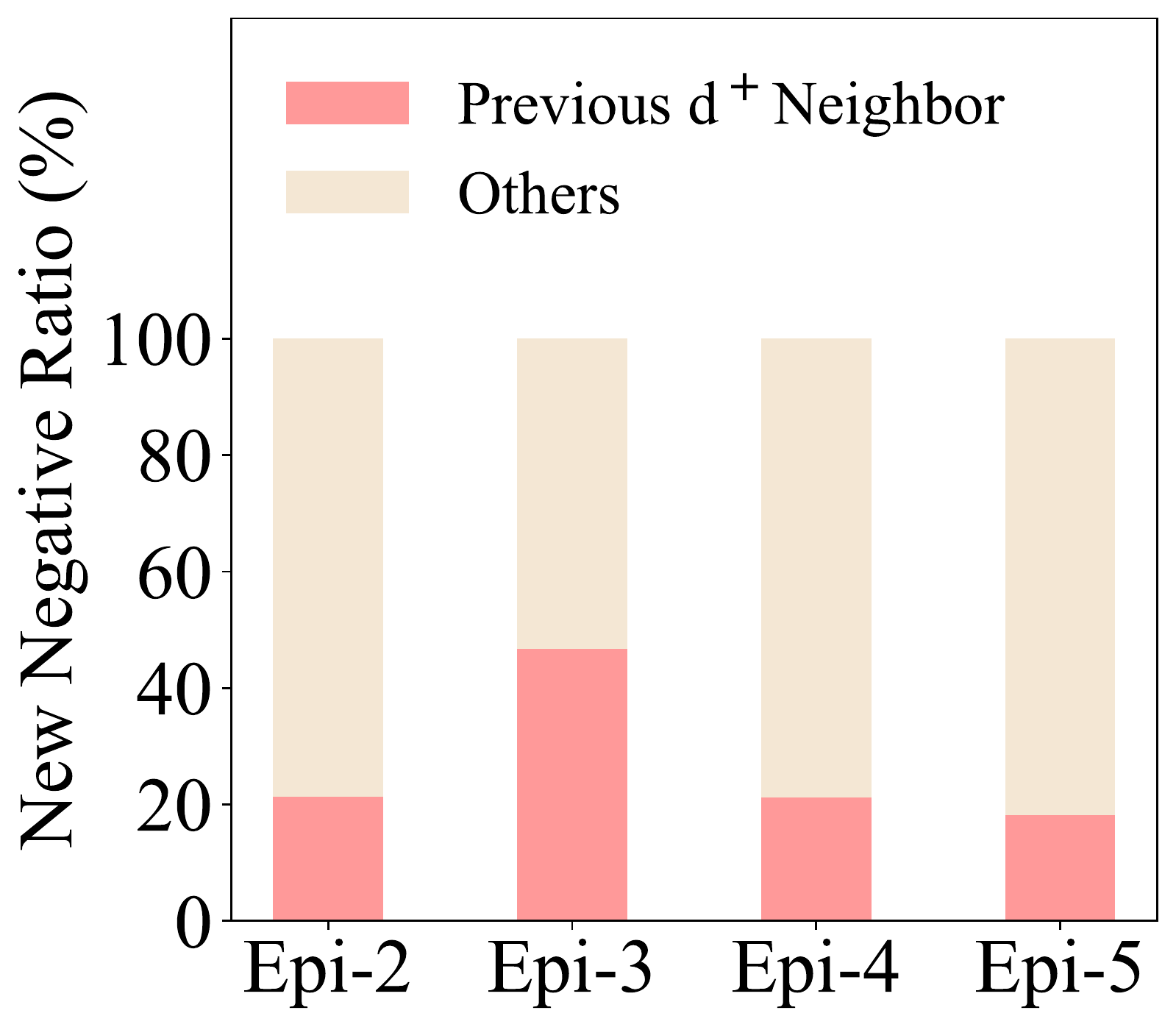}
        \centering
        \caption{ANCE (coCondenser).\label{fig:new_neg_frac}}
    \end{subfigure}
    \vspace{-0.2cm}
\caption{\label{fig:new_neg_compose} Composition of new negatives in each episode of ANCE with CyclicLR. New negatives are those first introduced in each episode.
}
\end{figure}
\begin{table*}[t]
\centering
\resizebox{\textwidth}{!}{
\begin{tabular}{l l l|l l|l l l|l l l}
\toprule
\multirow{2}{*}{\textbf{Methods}} 
& {\textbf{Parameters}} 
& {\textbf{Training Negatives}}
& \multicolumn{2}{c|}{\textbf{MS MARCO Dev}} 
& \multicolumn{3}{c|}{\textbf{Natural Question Test}} 
& \multicolumn{3}{c}{\textbf{TriviaQA Test}} \\
& (DR/Teacher) & (In DR Training) & MRR@10 & R@1K & R@5 & R@20 & R@100 & R@5 & R@20 & R@100 \\
\hline
 \multicolumn{3}{l|}{\textbf{Sparse Retrieval}}  & & & & & &  \\
BM25~\shortcite{yang2017anserini} & -- & -- & 18.7 & 85.7 & n.a & 59.1 & 73.7 & n.a & 66.9 & 76.7 \\

DeepCT~\shortcite{dai2019context} & -- & -- & 24.3 & 91.0 & n.a & n.a & n.a & n.a & n.a & n.a \\

docT5query~\shortcite{nogueira2019doc2query} & -- & -- & 27.7 & 94.7 & n.a & n.a & n.a & n.a & n.a & n.a \\ 

GAR~\shortcite{mao2021generation} & -- & -- & n.a. & n.a. & 60.9 & 74.4 & 85.3 & 73.1 & 80.4 & 85.7 \\
\hline
 \multicolumn{3}{l|}{\textbf{Dense Retrieval}}  & & & & & &  \\
 
DPR~\shortcite{dpr} & 110M & BM25 & 31.1 & 95.2 & n.a. & 78.4 & 85.4 & n.a. & 79.4 & 85.0 \\

DrBoost~\shortcite{lewis2021boosted} & 110M$\times$5 & Self (Boosted) & 34.4 & n.a. & n.a. & 80.9 & 87.6 & n.a & n.a & n.a \\

ANCE~\shortcite{ance} & 110M & BM25+Self & 33.0 & 95.9 & n.a. & 81.9 & 87.5 & n.a. & 80.3 & 85.3 \\

SEED-Encoder~\shortcite{lu2021less} & 110M & BM25+Self & 33.9 & 96.1 & n.a. & 83.1 & 88.7 & n.a. & n.a. & n.a. \\

RocketQA~\shortcite{rocketqa} & 110M & Self+Filter & 37.0 & 97.9 & 74.0 & 82.7 & 88.5 & n.a & n.a & n.a \\

ME-BERT~\shortcite{luan2021sparse} & 110M & BM25+Rand & 33.8 & n.a. & n.a & n.a & n.a & n.a & n.a & n.a \\

GTR-base~\shortcite{gtr} & 110M & RocketQA & 36.6 & 98.3 & n.a & n.a & n.a & n.a & n.a & n.a \\
Condenser~\shortcite{gao-callan-2021-condenser} & 110M & BM25+Self & 36.6 & 97.4 & n.a. & 83.2 & 88.4 & n.a. & 81.9 & 86.2 \\
coCondenser~\shortcite{cocondenser} & 110M & BM25+Self & 38.2 & \textbf{98.4} & 75.8 & 84.3 & 89.0 & 76.8 & 83.2 & \textbf{87.3} \\
coCondenser (Ours) & 110M & BM25+Self & 38.2 & \textbf{98.4} & 75.6 & 84.4 & 89.0 & 75.3 & 82.8 & 86.8 \\

\hline
\textbf{\model{}} & 110M & Self (Teleportation) & \textbf{39.1}$\text{}^{\sharp}$ & \textbf{98.4} & \textbf{77.0}$\text{}^{\sharp}$ & \textbf{84.9} & \textbf{89.7}$\text{}^{\sharp}$ & \textbf{76.9}$\text{}^{\sharp}$ & \textbf{83.4}$\text{}^{\sharp}$ & \textbf{87.3} \\ 
\hline
\hline

\multicolumn{5}{l|}{\textbf{For Reference:} Bigger models and/or distillation from reranking teachers} & \multicolumn{3}{l|}{} &
\\
DPR-PAQ-large~\shortcite{ouguz2021domain} & 355M & BM25+Self & 34.0 & n.a. & 76.9 & 84.7 & 89.2 & n.a & n.a & n.a \\
GTR-large~\shortcite{gtr} & 335M & RocketQA & 37.9 & 99.1 & n.a & n.a & n.a & n.a & n.a & n.a \\
GTR-XL~\shortcite{gtr} & 1.24B & RocketQA & 38.5 & 98.9 & n.a. & n.a & n.a & n.a & n.a & n.a \\
GTR-XXL~\shortcite{gtr} & 4.8B & RocketQA & 38.8 & {99.0} & n.a. & n.a & n.a & n.a & n.a & n.a \\
PAIR~\shortcite{ren2021pair} & 110M/330M & Pseudo Labels & 37.9 & 98.2 & 74.9 & 83.5 & 89.1 & n.a & n.a & n.a \\
RocketQA-v2~\shortcite{ren2021rocketqav2} & 110M/110M & RocketQA+Filter & 38.8 & 98.1 & 75.1 & 83.7 & 89.0 & n.a & n.a & n.a \\
AR2~\shortcite{zhang2021adversarial} & 110M/330M & BM25+Self & {39.5} & 98.6 & {77.9} & {86.0} & {90.1} & {78.2} & {84.4} & {87.9} \\
\bottomrule
\end{tabular}
}
\caption{\label{tab:overall} First stage retrieval performances. The total number of parameters used in the DR model and its teacher, if applicable, are listed under \textit{Parameters}. \textit{Training Negatives} list the negative training examples sampled from BM25, randomly, using DR model itself, or inherited from previous methods. DR systems using significantly more parameters than BERT$_\text{base}$ (110M) during training are listed \textit{For Reference} but not for fair comparisons. \textbf{$\sharp$} indicates statistically significant improvements over coCondenser (Ours).
} 
\end{table*}


\textbf{Lookahead negatives} predict the potential hard negatives in future training episodes using the neighbors of the positive document (Eqn.~\ref{eqn.future-neg}).
Information from future is known to be beneficial in many other scenarios, for example, to probe the optimization space~\citep{zhang2019lookahead} and to guide language generations~\citep{lu2021neurologic}.
However, obtaining future information often requires extra costly operation steps, e.g., actually performing an extra ANCE episode to collect.

Instead, we propose to use the neighbors of $d^+$ to approximate the negatives that may appear in future episodes, i.e., ``lookahead''. Intuitively, as the dense retrieval training is to pull closer $q$ and $d^+$ in the representation space, a side effect is that the neighbors around $d^+$ are also pulled to the query.
Figure~\ref{fig:new_neg_compose} shows that in ANCE training $d^+$ neighbors indeed contribute to a large fraction of new negatives introduced in future episodes, a handy feature for \model{} to efficiently incorporate future learning signals.

The sampling weights of momentum and lookahead negatives are controlled by hyperparameters ($\alpha, \beta$), which we simply set as $0.5$ without tuning.



\textit{Eliminating dependency on sparse retrieval.} 
As shown in the last section, the sparse retrieval negatives are dropped after the warm up episode.
\model{} thus directly starts from itself:
\begin{align}
&D_{0}^\text{Tele-neg} = \emptyset \\
&\theta_{0}^* = \text{Pretrained Weights}.
\end{align}
This removes the dependency of dense retrieval training on sparse retrieval.

\model{} introduces little computation overhead. 
The momentum negatives can be cached on disk and merged when constructing the training data for new episodes. The lookahead negatives need one extra ANN retrieval operation per positive document, which is efficient~\citep{faiss}. None of them increases GPU computations, which are the bottleneck in training.
Other than adding teleportation negatives, \model{} keeps other system components intact and can be plugged into with most dense retrieval systems.


\section{Experimental Methodology}
\label{sec:exp-method}

We describe our general experimental settings in this section and leave more details in Appendix.

\textbf{Dataset.} Following recent research, we conduct experiments on the first stage retrieval of three benchmarks: MS MARCO passage retrieval~\citep{bajaj2016ms}, Natural Questions~\citep{nq}, and TriviaQA~\citep{joshi2017triviaqa}. We use the exact same setting with DPR~\citep{dpr}, ANCE~\citep{ance}, and coCondenser~\citep{cocondenser}. More details of these datasets are in Appendix~\ref{app:dataset}.

\textbf{Our Methods.} \model{} fine-tunes on each benchmark starting from coCondenser~\citep{cocondenser}, a recent retrieval-oriented language model continuously pretrained from BERT$_\text{base}$.
The loss function is cross-entropy. We use 31 training negatives per query for MARCO, and 11 for NQ and TriviaQA, sampled from the union of Top 200 KNN results from query (ANCE), momentum, and lookahead negatives. Momentum and lookahead sample weights $(\alpha, \beta)$ are 0.5. 
All experiments are conducted at BERT$_\text{base}$ scale (110M parameters), the most popular setting currently.

\begin{table}[t]
\centering
\small
\begin{tabular}{l l l| c c}
\toprule
\textbf{$\alpha$} & \textbf{$\beta$} & \textbf{$N$}  & \textbf{MRR@10} & \textbf{R@1K} \\
\hline
0.5 & 0.5 & 47 & 39.1 & 98.3 \\
0.5 & 0.5 & 31 & 39.1 & 98.4 \\
0.5 & 0.5 & 23 & 38.9 & 98.3  \\
\hline
0.3 & 0.5 & 31 & 39.1 & 98.3 \\
0.7 & 0.5 & 31 & 39.0 & 98.4 \\
\hline
0.5 & 0.3 & 31 & 38.9 & 98.4 \\
0.5 & 0.7 & 31 & 38.9 & 98.3 \\
\bottomrule
  \end{tabular}
  \caption{\model{} on MARCO regarding to different hyperparameter values: the weights of momentum negatives $\alpha$, lookahead negatives $\beta$ and the total number of negative samples per $(q, d^+)$ pair $N$.\label{tab:hyperpara}}
\end{table}


\textbf{Baselines.} All previous first stage retrieval methods on the benchmarks are directly comparable with our performance empirically, as long as they follow the standard setting.
The fair baselines are those DR methods that are at the same BERT$_\text{base}$ scale, which we compare with their reported numbers.
We include our run of coCondenser for direct comparison, especially on TriviaQA where full implementation details were not publicly available. Descriptions of the baselines are in Appendix~\ref{app:basline}.

We also list the results from larger pretraining models and/or distillation from stronger cross-encoder reranking teachers, but only for reference purposes. How to more efficiently leverage the power of large scale pretraining models and how to scale techniques to billions of parameters are important future research directions.

\textbf{Implementation Details.} We implement \model{} using PyTorch~\cite{Paszke2019PyTorchAI} and HuggingFace~\cite{wolf-etal-2020-transformers}, and run all MARCO experiments on a single A100 GPU (40G) and all NQ and TriviaQA experiments on 4 V100 GPUs (32G). In each episode, the number of training epochs and query batch size is the same as in the previous work~\cite{cocondenser, dpr}. We directly use the last checkpoint in our run instead of selecting checkpoints based on empirical performances. 
The exact configurations of our experiments are listed in Appendix~\ref{app:exp} and our open-source repository.

\begin{table}[t]
\centering
\resizebox{\columnwidth}{!}{
\begin{tabular}{l l|l l}
\toprule
\textbf{Pretrain Models} &
\textbf{Methods} & \textbf{MRR@10} & \textbf{R@1K} \\
\hline
\multirow{6}{*}{BERT} & Zero-Shot & 0.09 & 0.21 \\
& DPR & 32.4 & 94.9 \\
& ANCE & 33.5 & 95.8 \\
\cline{2-4}
& Q-Neg & 28.7 & 88.3 \\
& Q-Neg w/ Mom-Neg & 35.6 & 96.0 \\
& Q-Neg w/ LA-Neg & 33.7 & 94.8 \\
& ANCE-Tele & \textbf{36.0}$\text{}^{\dagger \ddagger \flat \natural \S}$ & \textbf{96.2}$\text{}^{\dagger \ddagger \flat \natural \S}$ \\
\hline
\multirow{6}{*}{Condenser} & Zero-Shot & 0.61 & 11.4 \\
& DPR & 33.8 & 96.1 \\
& ANCE & 35.0 & 96.7 \\
\cline{2-4}
& Q-Neg & 30.4 & 87.8 \\
& Q-Neg w/ Mom-Neg & 37.0 & 97.0 \\
& Q-Neg w/ LA-Neg & 36.7 & 96.9 \\
& ANCE-Tele & \textbf{37.2}$\text{}^{\dagger \ddagger \flat \natural \S}$ & \textbf{97.1}$\text{}^{\dagger \ddagger \flat \natural}$ \\
\hline
\multirow{6}{*}{coCondenser} & Zero-Shot & 11.4 & 77.1 \\
& DPR & 36.2 & 97.7 \\
& ANCE & 36.8 & 98.1 \\
\cline{2-4}
& Q-Neg & 36.6 & 95.3 \\
& Q-Neg w/ Mom-Neg & 38.6 & \textbf{98.4} \\
& Q-Neg w/ LA-Neg & 38.8 & 98.3 \\
& ANCE-Tele & \textbf{39.1}$\text{}^{\dagger \ddagger \flat \natural}$ & \textbf{98.4}$\text{}^{\dagger \ddagger \flat \natural}$ \\
\bottomrule
  \end{tabular}
}
  \caption{\label{tab:ablation} MARCO performances with different pretrained models and training negative samples. All methods are trained for three episodes, except for Zero-Shot and DPR. Superscripts indicate statistically significant improvements over Zero-Shot$\text{}^{\dagger}$, DPR$\text{}^{\ddagger}$, ANCE$\text{}^{\flat}$, Query-Neg$\text{}^{\natural}$, Query-Neg w/ Momentum-Neg$\text{}^{\sharp}$, Query-Neg w/ Lookahead-Neg$\text{}^{\S}$.}
\end{table}


\section{Evaluation Results}
\label{sec:eval-res}

This section first evaluates \model{} and its ablations. 
Then we analyze the influences and characteristics of teleportation negatives.

\begin{figure*}[t]
  \centering
    \vspace{-0.1cm}
      \begin{subfigure}[t]{0.24\linewidth}
        \includegraphics[height=3.2cm]{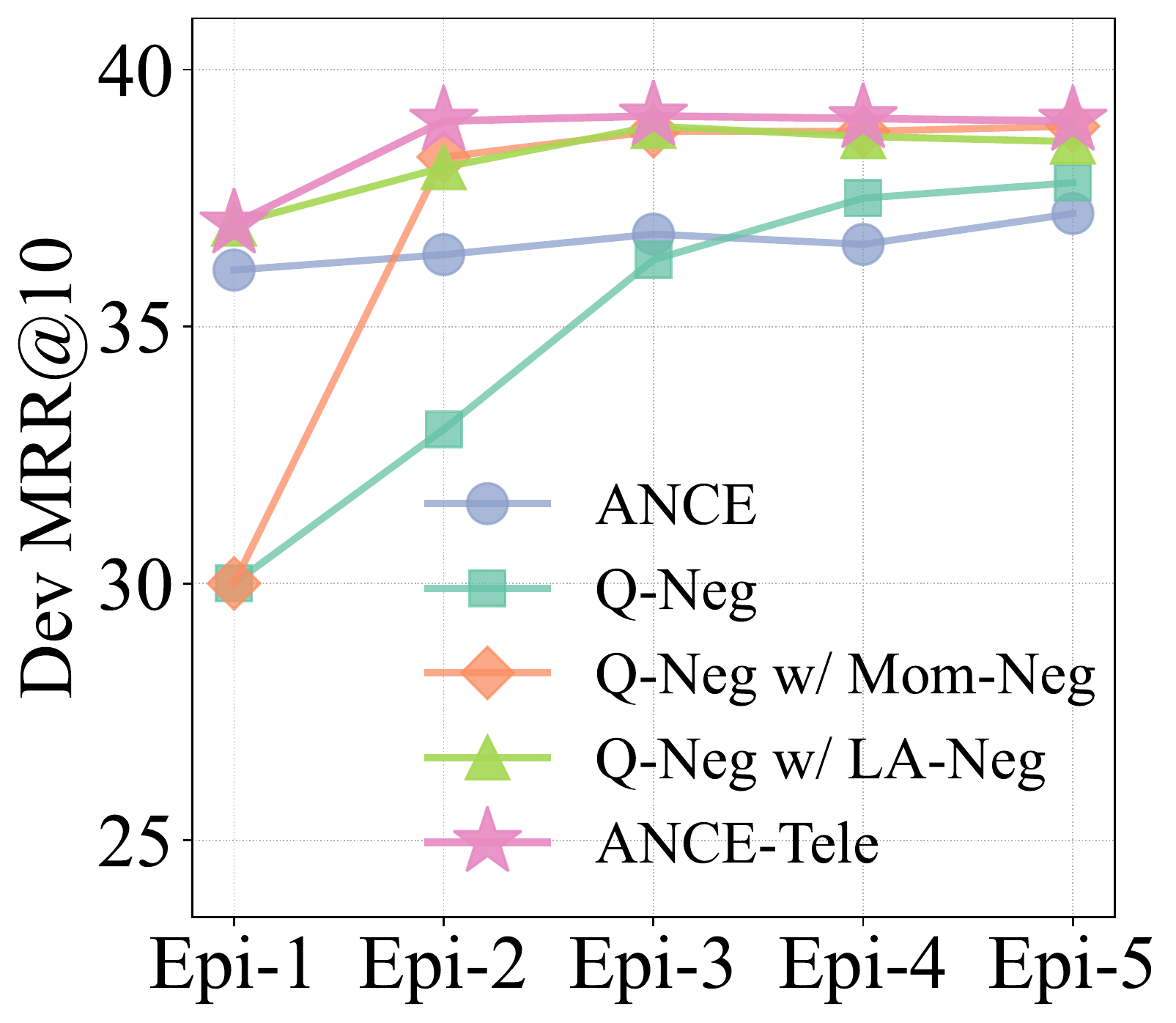}
        \caption{Iteration performance.\label{fig:coverge_speed_ite}}
    \end{subfigure}
    \begin{subfigure}[t]{0.24\textwidth}
        \includegraphics[height=3.2cm]{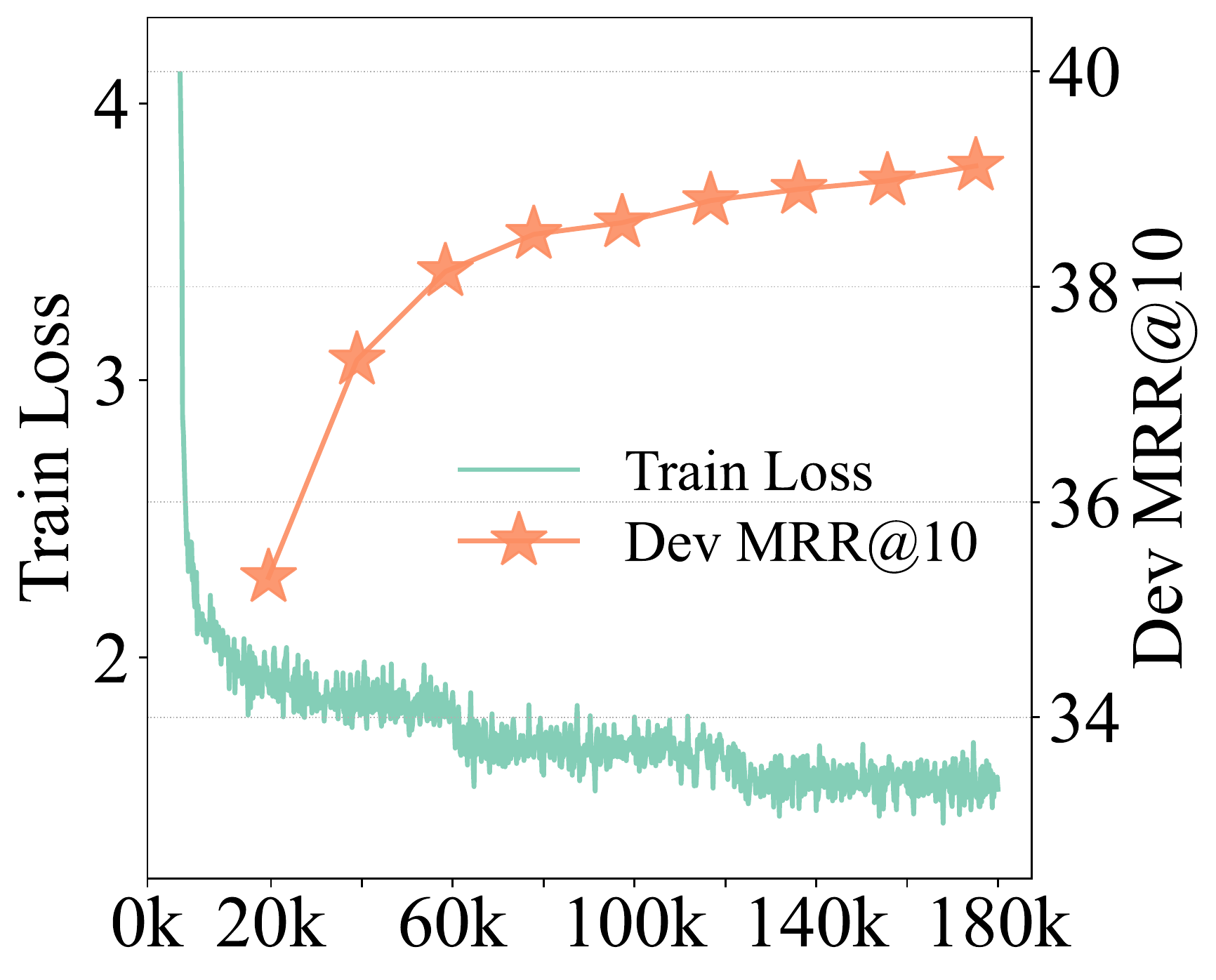}
        \caption{ANCE-Tele (Epi-3). \label{fig:tele-train-loss}}
    \end{subfigure}
    \hspace{0.25cm}
    \begin{subfigure}[t]{0.24\linewidth}
        \includegraphics[height=3.2cm]{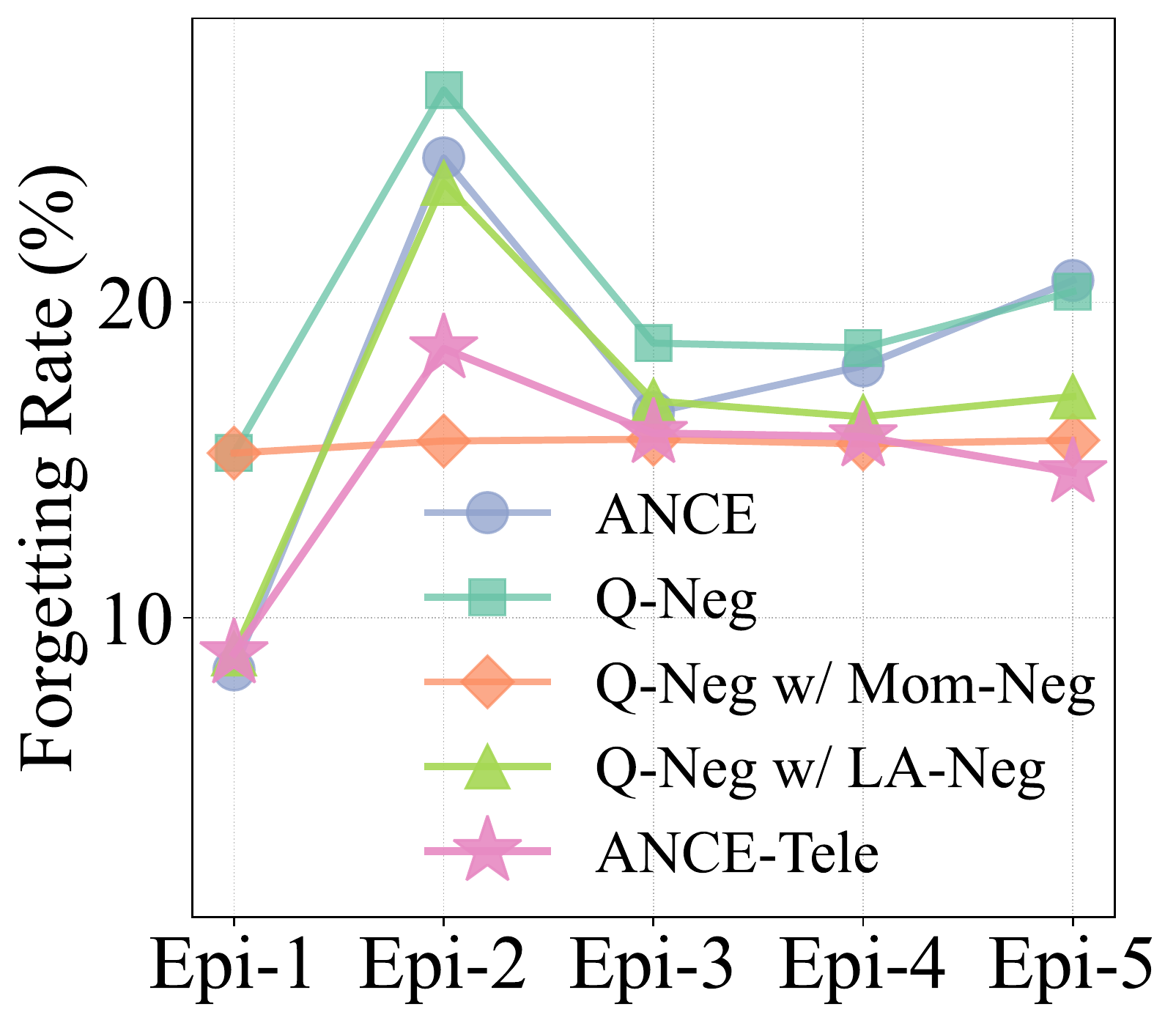}
        \caption{Forgetting Rate.\label{fig:mom_ana_forgot}}
    \end{subfigure}
     \hspace{-0.4cm}
    \begin{subfigure}[t]{0.24\linewidth}
        \includegraphics[height=3.4cm]{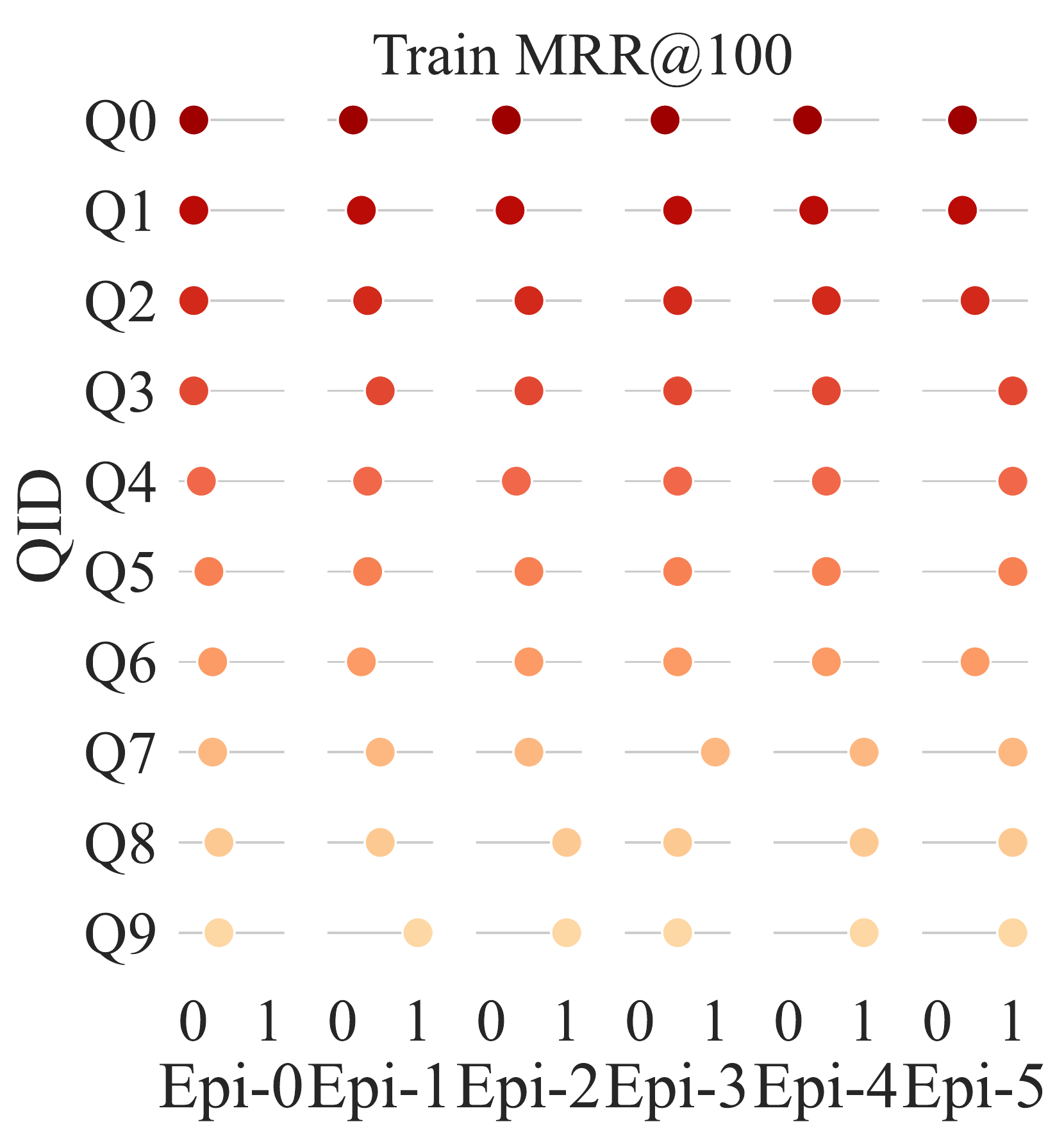}
        \centering
        \caption{Per Query Accuracy. \label{fig:tele-train-mrr}}
    \end{subfigure}
    \hspace{-0.4cm}
    \vspace{-0.1cm}
\caption{Training behavior with \model{} variants. (a) shows the iteration performance. (b) presents the training loss and dev curve of ANCE-Tele at Epi-3. (c) exhibits the forgetting rate. (b) shows the accuracy of ten training queries during ANCE-Tele training, which are the same queries sampled in the previous ANCE analysis. \label{fig:epi_ana}}
\end{figure*}


\subsection{Overall Results}
\label{sec:overal}

The overall performances are listed in Table~\ref{tab:overall}. 
At BERT$_\text{base}$ size,  \model{} outperforms previous state-of-the-arts on nearly all metrics. As shown in \textit{Training Negatives}, previous methods use various ways to construct training signals: sparse retrieval negatives, self negatives, and/or additional filters. \model{} only uses teleportation negatives from itself and does not depend on sparse retrieval.

\model{} shows competitive performances with systems with more parameters and/or distillations from reranking teachers.
It outperforms GTR-XXL on MRR@10 at MARCO, albeit the latter uses T5-XXL~\citep{gtr} with about 50x more parameters.
\model{} also achieves similar performance with AR2, which jointly trains retriever with knowledge distilled from an ERNIE$_\text{Large}$ reranking model~\citep{zhang2021adversarial}.

\subsection{Ablations}
\label{sec:ablation}



We perform two ablations to study the hyper-parameters and design choices of \model{}. 

\textbf{Hyperparameters.} We keep the design choices of \model{} as simple as possible. The only hyperparameters introduced are the $\alpha$ and $\beta$ to balance momentum and lookahead negatives. Another notable hyperparameter is the number of negatives sampled per $(q, d^+)$ pair,  inherited from ANCE. In Table~\ref{tab:hyperpara} we show that \model{} is robust to these hyperparameter variations.

\textbf{Negative Sampling.} Table~\ref{tab:ablation} lists the performance of \model{} when starting from different pretrained models with different negatives. 
We use Condenser as representatives of pretraining models using information bottleneck for better sequence embedding~\citep{gao-callan-2021-condenser, lu2021less, wang2021tsdae}, and coCondenser to represent pretraining models with sequence contrastive learning~\citep{meng2021coco, gtr, cocondenser}. The baseline performances of zero-shot, DPR (BM25 negatives), and ANCE with Condenser and coCondenser confirm the benefits of these two pretraining techniques. 

\model{} provides robust improvements despite different pretrained starting points.
Among different negative selection approaches, Query negatives alone lag behind ANCE after three training episodes, confirming the important role of BM25 negatives in ANCE. Adding either momentum or lookahead negatives eliminates the dependency on sparse retrieval, even when starting from BERT whose zero-shot performance is barely nonrandom.
The two provide a significant boost when added individually or combined. 
The benefits are further analyzed in next experiments.

\begin{figure*}[t]
\centering 
    \hspace{-0.2cm}
    \vspace{-0.1cm}
        \begin{subfigure}[t]{0.24\linewidth}
        \includegraphics[height=3.2cm]{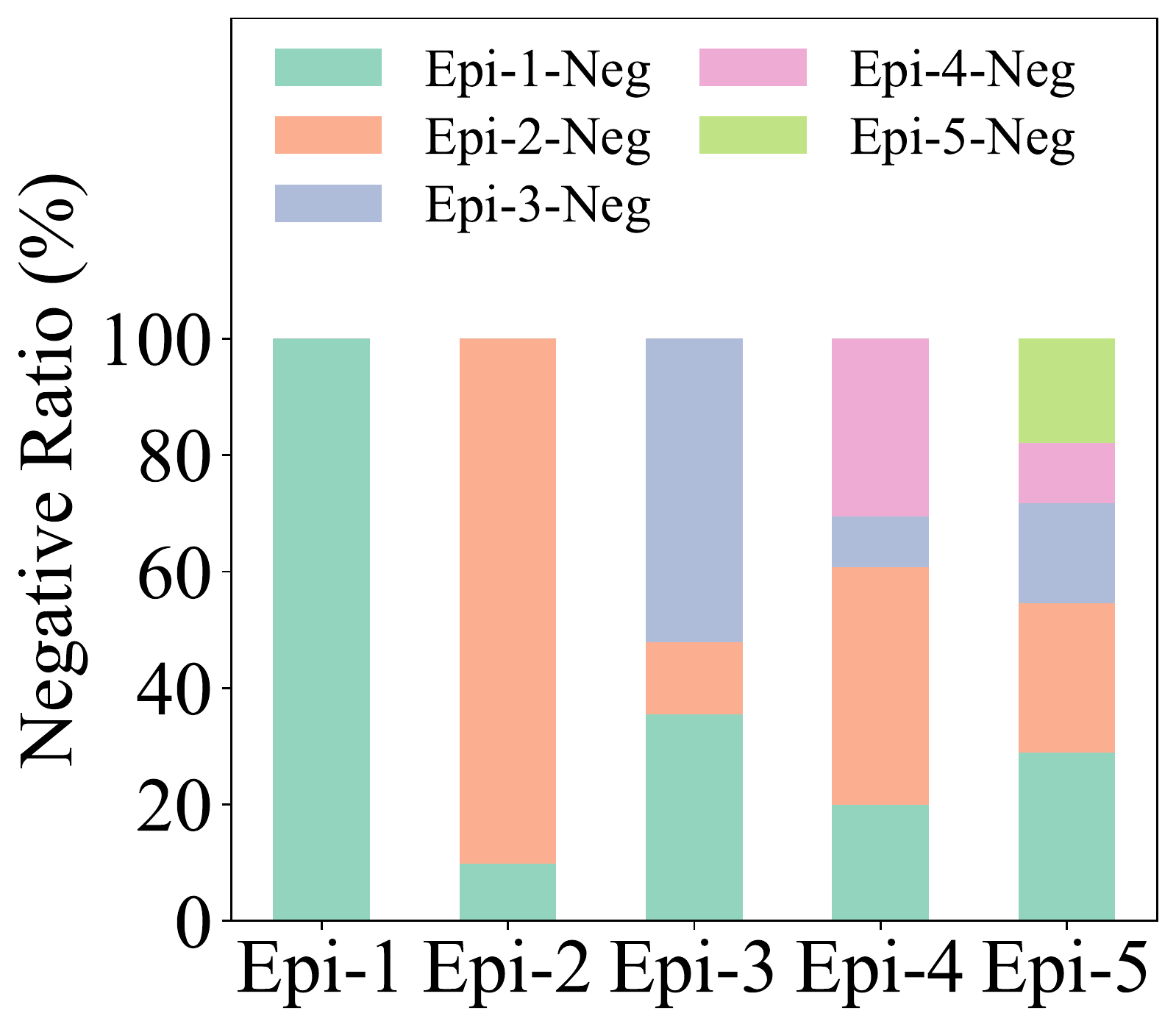}
        \centering
        \caption{Q-Neg.\label{fig:qry-neg-dist}}
    \end{subfigure}
    \begin{subfigure}[t]{0.24\linewidth}
        \includegraphics[height=3.2cm]{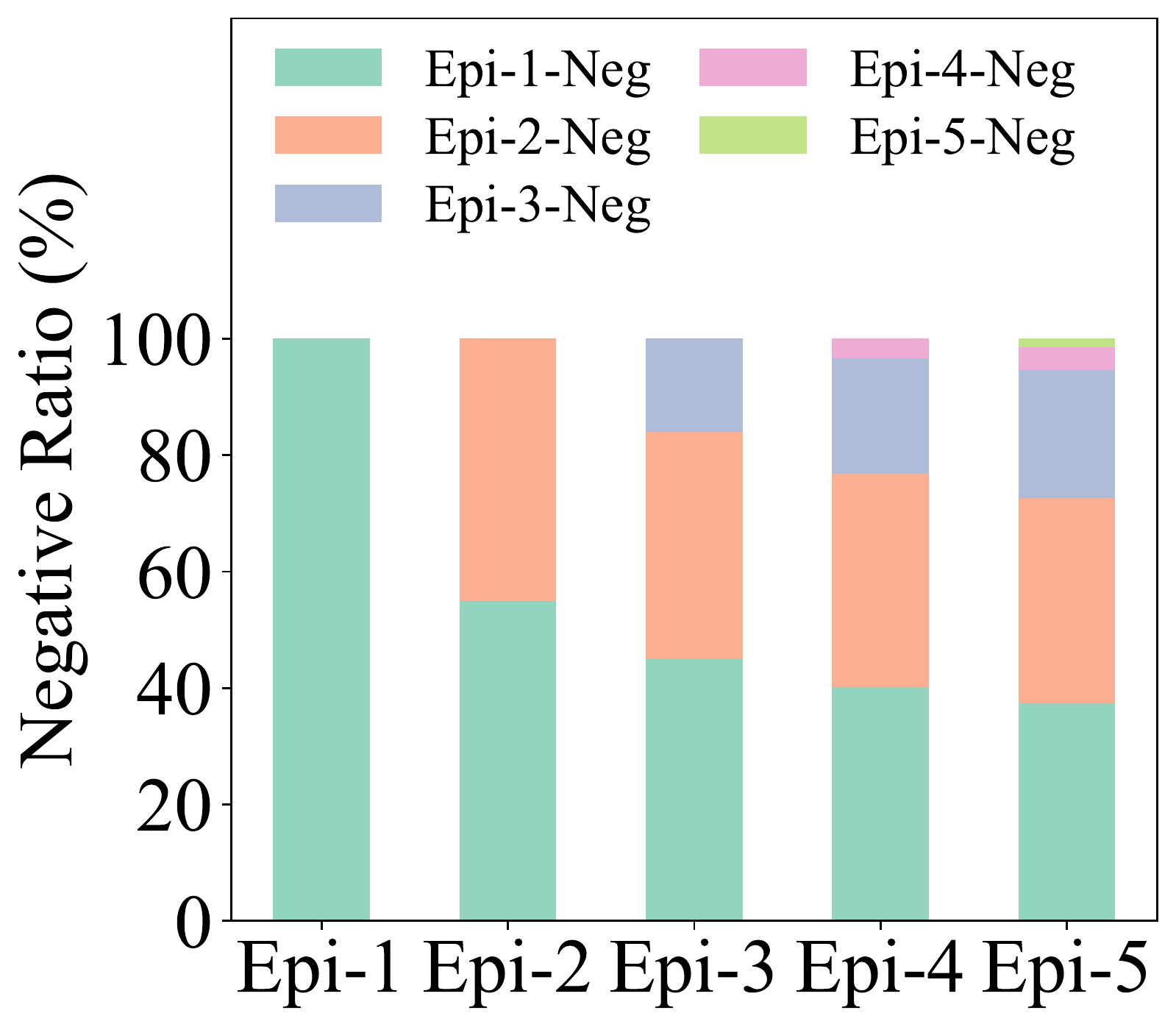}
        \centering
        \caption{Q-Neg w/ Momentum.\label{fig:mom-neg-dist}}
    \end{subfigure}
    \hspace{0.2cm}
    \begin{subfigure}[t]{0.24\linewidth}
        \includegraphics[height=3.2cm]{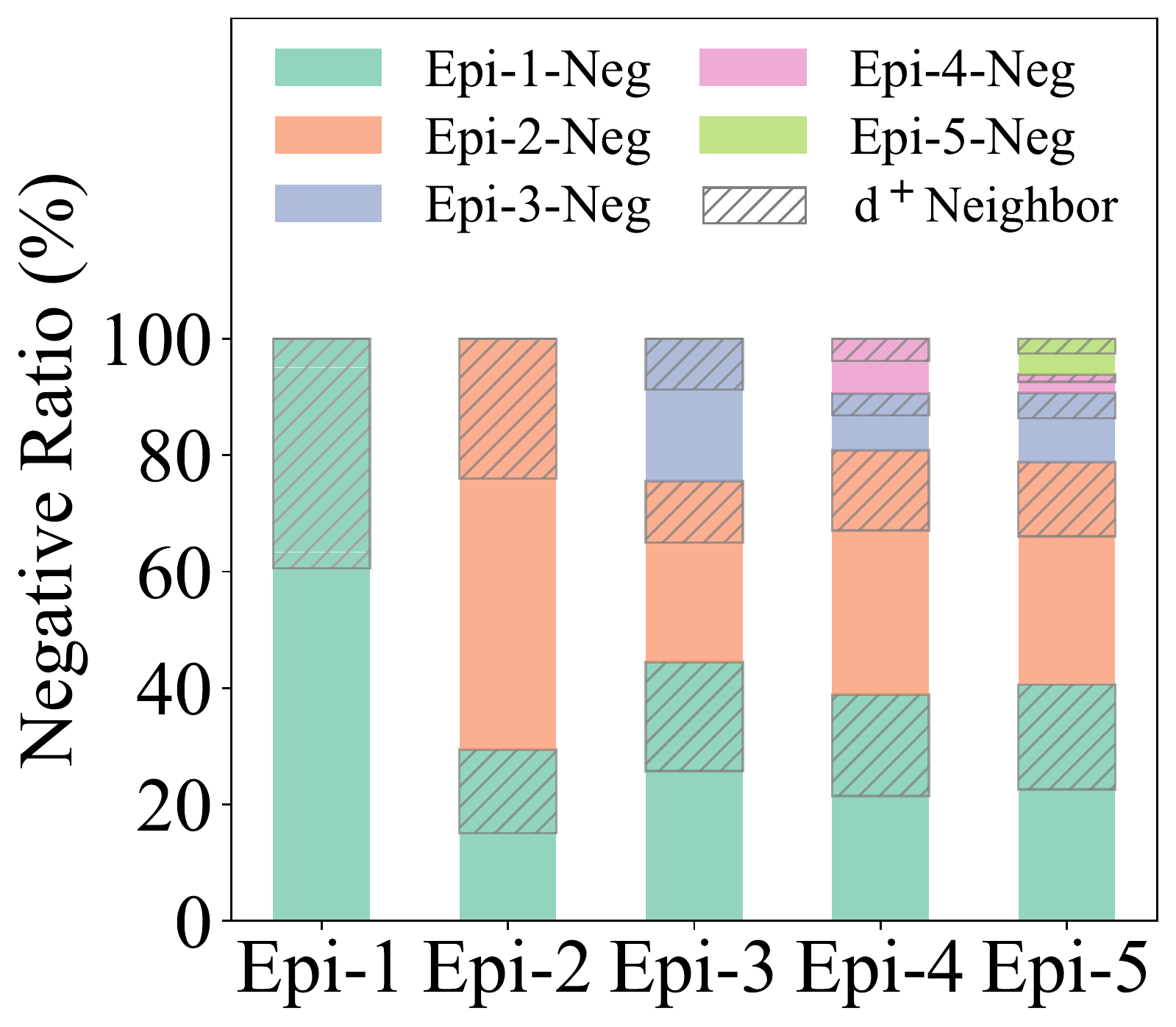}
        \caption{Q-Neg w/ Lookahead. \label{fig:la-neg-dist}}
    \end{subfigure}
    \hspace{-0.1cm}
        \begin{subfigure}[t]{0.24\textwidth}
        \includegraphics[height=3.2cm]{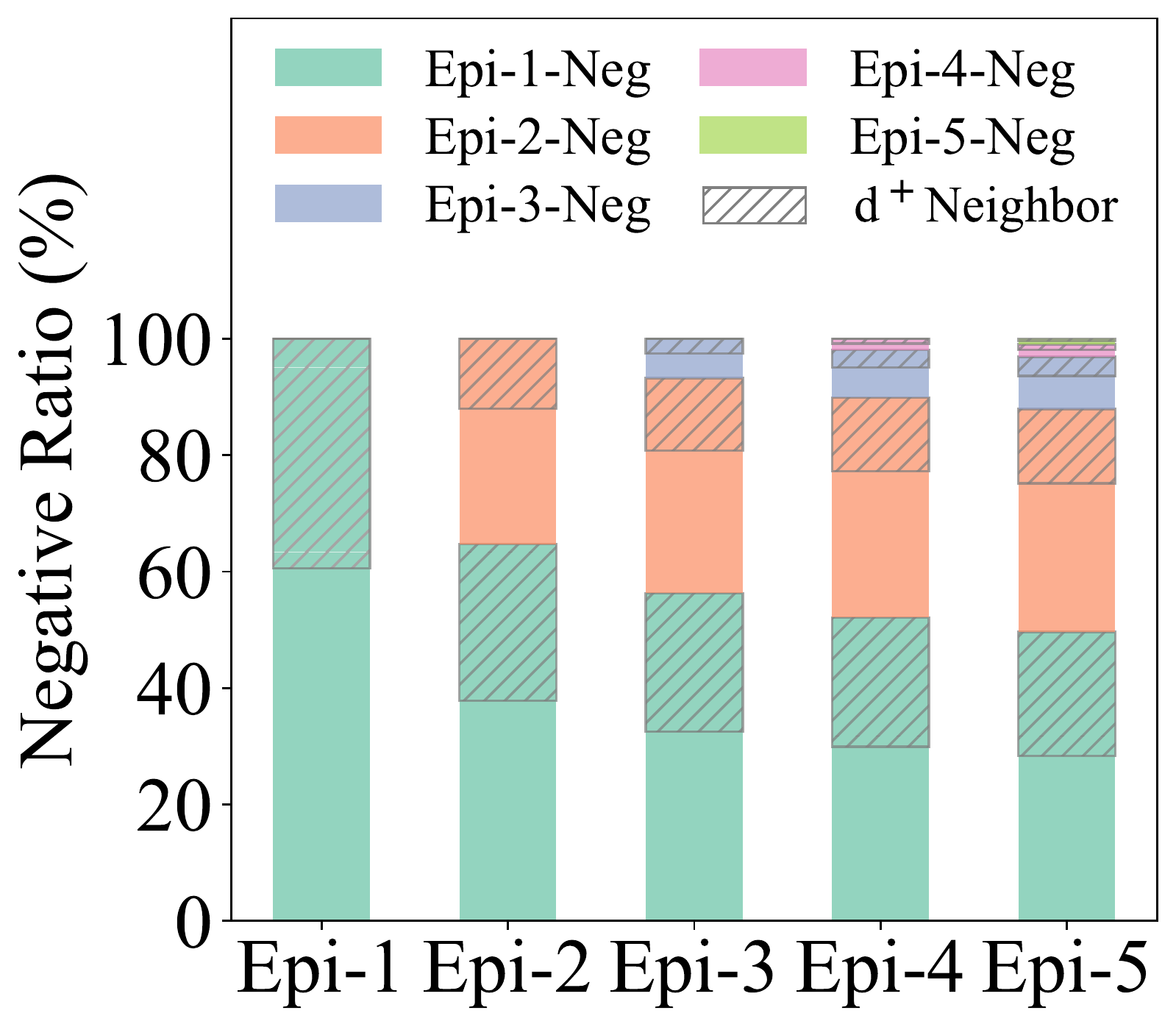}
        \centering
        \caption{\model{}. \label{fig:tele-neg-dist}}
    \end{subfigure}
    \hspace{-0.1cm}
    \vspace{-0.1cm}
\caption{\label{fig:Negative Variance} Composition of training negatives in \model{} variations. Only lookahead includes $d^+$ neighbors.
}
\end{figure*}


\subsection{Influence of Teleportation Negatives}

In this experiment we evaluate the influence of teleportation negatives in the training process.

\textbf{Training Stability.} In Figure~\ref{fig:coverge_speed_ite} we plot the performance of \model{} after different training episodes. 
In comparison to ANCE in Figure~\ref{fig:ance_perform}, \model{} variants with momentum negatives are more stable.
Lookahead negatives improve converging speed. Adding them improves accuracy at the first episode by more than 20\% relatively, a key contributor to the elimination of sparse retrieval negatives. We also show experimentally that adding BM25 negatives brings no additional benefits to ANCE-Tele, as presented in Table~\ref{app-tab:ance-tele-variants} (Appendix~\ref{app:tele-variants}). In addition, Figure~\ref{fig:tele-train-loss} zooms into the third episode of ANCE-Tele and further confirms its stability and converging speed.




\textbf{Forgetting Rate.} Figure~\ref{fig:mom_ana_forgot} plot the forgetting ratio of \model{} variations. 
Q-Neg alone yields large forgetting rate, especially at the second episode  (Figure~\ref{fig:coverge_speed_ite}).
The momentum-negatives significantly reduce the forgetting rate throughout training. Including past negatives does remind the model to maintain learned knowledge. Furthermore, we revisit the ten training queries analyzed in Figure~\ref{fig:forget-example} and track the performance of ANCE-Tele on them. As shown in Figure~\ref{fig:tele-train-mrr}, the per query MRR oscillation reduces significantly in ANCE-Tele. Most of them maintain high accuracy throughout training episodes.

\subsection{Composition of Teleportation Negatives}

This experiment analyzes the composition of teleportation negatives. The sources of training negatives (by first encounter) are plotted in Figure~\ref{fig:Negative Variance}.

Using Q-Neg alone suffers from negative swing (Figure~\ref{fig:qry-neg-dist}). Many negatives selected in Epi-1 got pushed away in Epi-2 (\textit{ratio} $\downarrow$) and then reemerge in Epi-3 (\textit{ratio} $\uparrow$). It corresponds to the high forgetting rate at the end of Epi-2 (Figure~\ref{fig:mom_ana_forgot}).
Momentum negatives nearly eliminate this issue via a smooth combination of new and inherited negatives. In Figure~\ref{fig:mom-neg-dist}, hard negatives are introduced gradually and remain in effect for multiple episodes.

The neighbors of positive document (lookahead negatives) is a decent approximation of oracle future negatives in later episodes. In Figure~\ref{fig:la-neg-dist}, significantly lower fractions of new negatives are introduced in later episodes, compared to Q-Neg. The $d^+$ neighbors from the last episode cover 20\%-30\% of would-be new negatives of the current episode, pretty efficient given its negligible computing cost. 

As shown in Figure~\ref{fig:tele-neg-dist}, the teleportation negatives are composed by a diverse collection of negative sources and evolve more smoothly through episodes. This helps \model{} improve the optimization stability, efficiency, and reduces catastrophic forgetting in dense retrieval training. Appendix~\ref{app:tele-variants} also studies the influence of removing the overlapping part from multiple negative sources.

Appendix~\ref{app:case} exhibits some example negatives which ANCE learned, forgot, and then relearned.
The swing between different distractor groups aligns well with our intuitions in information retrieval. Those irrelevant documents cover different partial meanings of the query and require different ways to detect.
\model{} eliminates this behavior by combining them via teleportation negatives.

\section{Conclusions}
\label{sec:conclusion}

We present \model{}, a simple approach that effectively improves the stability of dense retrieval training. Our investigation reveals the issues underlying the training instability:
the catastrophic forgetting and negative swing behaviors. \model{} resolves these issues by introducing teleportation negatives, which smooth out the learning process with momentum and lookahead negatives.


\model{} leads to strong empirical performance on web search and OpenQA with improved training stability,  convergence speed, and reduced catastrophic forgetting.
Our analysis demonstrates the benefits of teleportation negatives and their behavior during training iterations.
Overall, \model{} addresses an intrinsic challenge in dense retrieval training with reduced engineering effort and minimum computation overhead. It can be used as a plug-in upgrade for the first stage retrieval of many language systems.

\section*{Limitations}




One limitation of \model{} is the lack of more detailed characterization of the representation space, in dense retrieval and other embedding-based matching tasks. Better understanding of the representation space will enable the development of more automatic ways to navigate the training signal space. The training signal selection is getting more and more important with deep neural systems. We need more tools to capture, analyze, and improve this new aspect of data-driven AI.

Though \model{} is robust to different pretraining models, whether tailored to retrieval or not, there is still limited understanding on the relationship between the pretraining task and the pretrained model's generalization ability in dense retrieval. Recent research observed several mismatches between the two, but we still do not quite fully understand their interactions.


\section*{Acknowledgments}
This work is partly supported by Institute Guo Qiang at Tsinghua University and NExT++ project from the National Research Foundation, Prime Minister’s Office, Singapore under its IRC@Singapore Funding Initiative. We thank all anonymous reviewers for their suggestions.

\bibliography{citation_ready}
\bibliographystyle{acl_natbib}

\clearpage
\appendix

\section{Dataset Details}
\label{app:dataset}

In our experiments, we utilize three evaluation datasets, i.e., MS MARCO, NQ, and TriviaQA, and their statistical details are shown in Table~\ref{app-tab:dataset}.

\begin{itemize}
    \item \textbf{MS MARCO}: The MS MARCO Passage Ranking~\cite{bajaj2016ms} is a large scale of web search dataset. The queries are constructed from Bing’s query logs, and each query comes with at least one related passage.
    
    \item \textbf{NQ}: Natural Questions~\cite{nq} is a widely used question-answering dataset constructed on Wikipedia. The questions come from the Google search engine, and the answers are identified as text spans in the Wikipedia article.
    
    \item \textbf{TriviaQA}: The TriviaQA~\cite{joshi2017triviaqa} is a reading comprehension dataset collected from Wikipedia and the web. Trivia enthusiasts author the question-answer pairs.
\end{itemize}

Following the prior work~\cite{ance}, we report MRR@10 and Recall@1K on the Dev set of MARCO. The Mean Reciprocal Rank (MRR) is the average of the inverse ranking positions of the first retrieved relevant passages for all queries. Recall@1K represents the proportion of queries containing relevant passages in the top 1K retrieved passages. We report Recall@\{5,10,100\} on the test set of NQ and TriviaQA, and use passage titles on these three datasets, which is consistent with the previous research~\cite{dpr, cocondenser}. Statistical significance is examined by permutation test with $p<0.05$.



\section{Main Baselines}
\label{app:basline}


This section briefly introduces the DR methods in Table~\ref{tab:overall}, the main baselines of \model{}. 

\textbf{DPR}~\cite{dpr} is a classical dual-encoder DR method, which maps the query and passage to dense vectors separately, using the \texttt{[CLS]} token from the pre-trained model. Based on the dense vectors, the dot product is used to compute their similarity. DPR is trained with in-batch and BM25 negatives, using BERT$_\text{base}$ as initial models.
    

\textbf{DrBoost}~\cite{lewis2021boosted} is an ensemble-like DR method that utilizes negatives sampled from itself for iterative training. The training starts with BERT$_\text{base}$ and random negatives. Its final representation of the query and passage is the concatenation of the output dense vectors of all component models during training.


\textbf{ANCE}~\cite{ance} is a popular DR training strategy. It starts warm-up training based on BM25 negatives and continues iterative training using the hard negatives retrieved from the latest checkpoint. RoBERTa$_\text{base}$ is the initial model.
    
\textbf{SEED-Encoder}~\cite{lu2021less} is an IR-oriented pre-trained model with an auto-encoder architecture.
Pretraining configures the encoder with a relatively weak decoder to push the encoder to obtain more robust text representations. Fine-tuning only leverages the encoder, whose parameter quantity is equivalent to BERT$_\text{base}$ and the fine-tuning process is the same as ANCE.

\begin{table}[t]
\centering
\small
\resizebox{\columnwidth}{!}{
\begin{tabular}{l c c c c}
\toprule
\textbf{Datasets} & \textbf{Train}  & \textbf{Dev} & \textbf{Test} & \textbf{Corpus Size} \\
\hline
MS MARCO & 502,939 & 6,980 & 6,837 & 8,841,823 \\
NQ & 79,168 & 8,757 & 3,610 & 21,015,324 \\
TriviaQA & 78,785 & 8,837 & 11,313 & 21,015,324 \\
\bottomrule
  \end{tabular}
}
\caption{The statistics of evaluation dataset. Train denotes the original training examples without filtering. \label{app-tab:dataset}} 
\end{table}



\begin{table*}[t]
\centering
\small
\begin{tabular}{l|lll}
\toprule
\textbf{Hyperparameters} & \textbf{MS MARCO}  & \textbf{NQ} & \textbf{TriviaQA} \\
\hline
Max query length & 32 & 32 & 32 \\
Max passage length & 128 & 156 & 156 \\
Negative mining depth & 200 & 200 & 200 \\
Batch size (query size per batch) & 8 & 128 & 128 \\
Positive number per query & 1 & 1 & 1 \\
Negative number per query & 31 & 11 & 11 \\
Initial model & \texttt{co-condenser-marco} & \texttt{co-condenser-wiki} & \texttt{co-condenser-wiki} \\
Learning rate & 5e-6 & 5e-6 & 5e-6 \\ 
Optimizer & AdamW & AdamW & AdamW \\
Scheduler & Linear & Linear  & Linear \\
Warmup ratio & 0.1 & 0.1 & 0.1 \\
Training epoch & 3 & 40 & 40 \\
Momentum negative weight $\alpha$ & 0.5 & 0.5 & 0.5 \\
Lookahead negative weight $\beta$ & 0.5 & 0.5 & 0.5 \\
\hline
Epi-1 new negative mining source & \texttt{co-condenser-marco} & \texttt{co-condenser-wiki} & \texttt{co-condenser-wiki} \\
\hline
Epi-2 new negative mining source & Epi-1 (20k step) & Epi-1 (2k step) & Epi-1 (2k step) \\

\hline
Epi-3 new negative mining source & Epi-2 (20k step) & Epi-2 (2k step) & Epi-2 (2k step) \\
\bottomrule
  \end{tabular}
  \caption{Hyperparameters of \model{} training. ANCE-Tele uses coCondenser as the initial pre-trained model. We use the \texttt{co-condenser-marco} version on MARCO, which is continuously pre-trained on the MARCO corpus. On NQ and TriviaQA, we use the \texttt{co-condenser-wiki} version, which continues pre-training on the Wikipedia corpus. We directly utilize the open-source model files of \texttt{co-condenser-marco} and \texttt{co-condenser-wiki} from the huggingface.co community. \label{app-tab:hyperpara}}
\end{table*}

\textbf{RocketQA}~\cite{rocketqa} is a dual-encoder DR initialized with the pre-trained model ERNIE 2.0 (Base). The training uses cross-batch negatives and a re-ranker to filter the false negatives. Besides, it is augmented with additional training data.

\textbf{ME-BERT}~\cite{luan2021sparse} represents each passage with multiple vectors and down-projects the vectors to the final representation. The training leverages random, BM25, and in-batch negatives, and utilizes BERT$_\text{base}$ for initialization.

\textbf{GTR-base}~\cite{gtr} is a dual-encoder DR method, initialized using the encoder part of the pre-trained model T5$_\text{base}$, trained with hard negatives released by RocketQA~\cite{rocketqa}.

\textbf{Condenser}~\cite{gao-callan-2021-condenser} is also an IR-oriented pre-trained model based on BERT$_\text{base}$.
During pretraining, it enhances the representation ability of \texttt{[CLS]} token by changing the connections between different layers of Transformer blocks. Fine-tuning uses BM25 and self-mining negatives.

\textbf{coCondenser}~\cite{cocondenser} adds a contrastive pre-training task to Condenser. The task randomly extracts spans from the specific corpus, regarding spans from the same/different passages as positive/negative examples, and learns to discriminate them. The model scale and fine-tuning process are the same as that of Condenser.

\begin{table}[t]
\centering
\small
\begin{tabular}{llll}
\toprule
\textbf{Stage} & \textbf{MS MARCO}  & \textbf{NQ} & \textbf{TriviaQA} \\
\hline
Epi-1 & 2.5h & 1.2h & 1.2h \\
Epi-2 & 2.5h & 1.2h & 1.2h \\
Epi-3 & 23.5h & 10.8h & 10.8h \\
Index refresh & 1.2h & 2.7h & 2.7h \\
Refresh number & 3 & 3 & 3 \\
\hline
Overall & 32.1h & 21.3h & 21.3h \\
\bottomrule
  \end{tabular}
  \caption{Training time for \model{} with three training episodes. Training on MARCO uses a single A100 GPU (40G), and training on NQ and TriviaQA utilizes 4 V100 GPUs (32G).\label{app-tab:train-time}}
\end{table}

\begin{table}[t]
\centering
\small
\begin{tabular}{l l}
\toprule
\textbf{Hyperparameters} & \textbf{Standard ANCE} \\
\hline
Negative refresh step & 10k \\ 
Negative mining depth & 200 \\
Batch size (query size per batch) & 8 \\
Positive number per query & 1 \\
Negative number per query & 7 \\
Learnig rate & 1e-6 \\
Optimizer & AdamW \\
Scheduler & Linear \\
Warmup step & 5k \\
Cyclical warmup & w/o \\
\bottomrule
  \end{tabular}
\caption{Hyperparameters of training standard ANCE in the analysis of Section~\ref{sec:learn-instable}.\label{app-tab:ance-param}} 
\end{table}



\begin{figure*}[t]
\centering 
    \begin{subfigure}[t]{0.24\linewidth}
        \includegraphics[height=3.2cm]{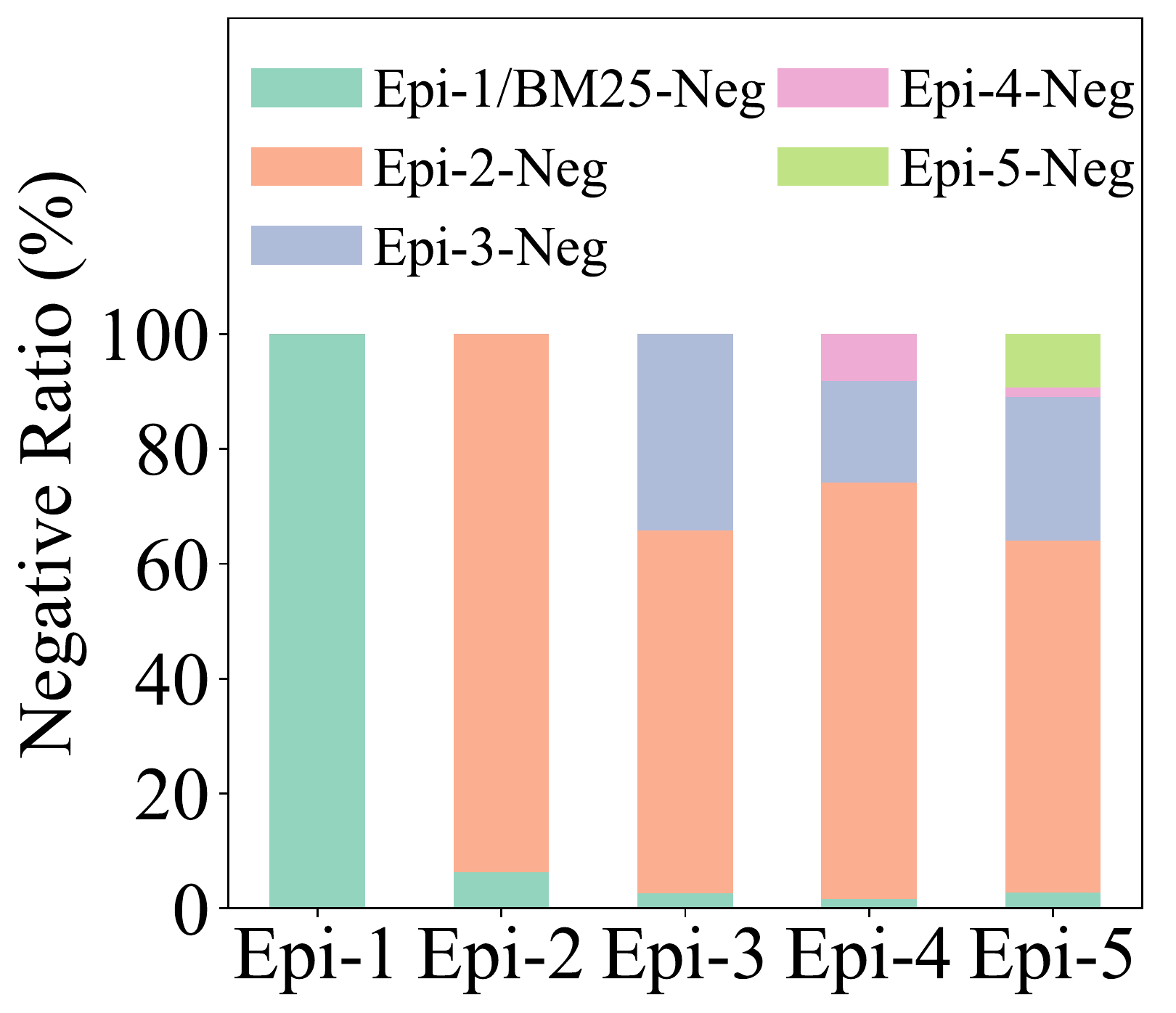}
        \caption{Standard/BERT.\label{app-fig:stand-ance-bert}}
    \end{subfigure}
        \begin{subfigure}[t]{0.24\linewidth}
        \includegraphics[height=3.2cm]{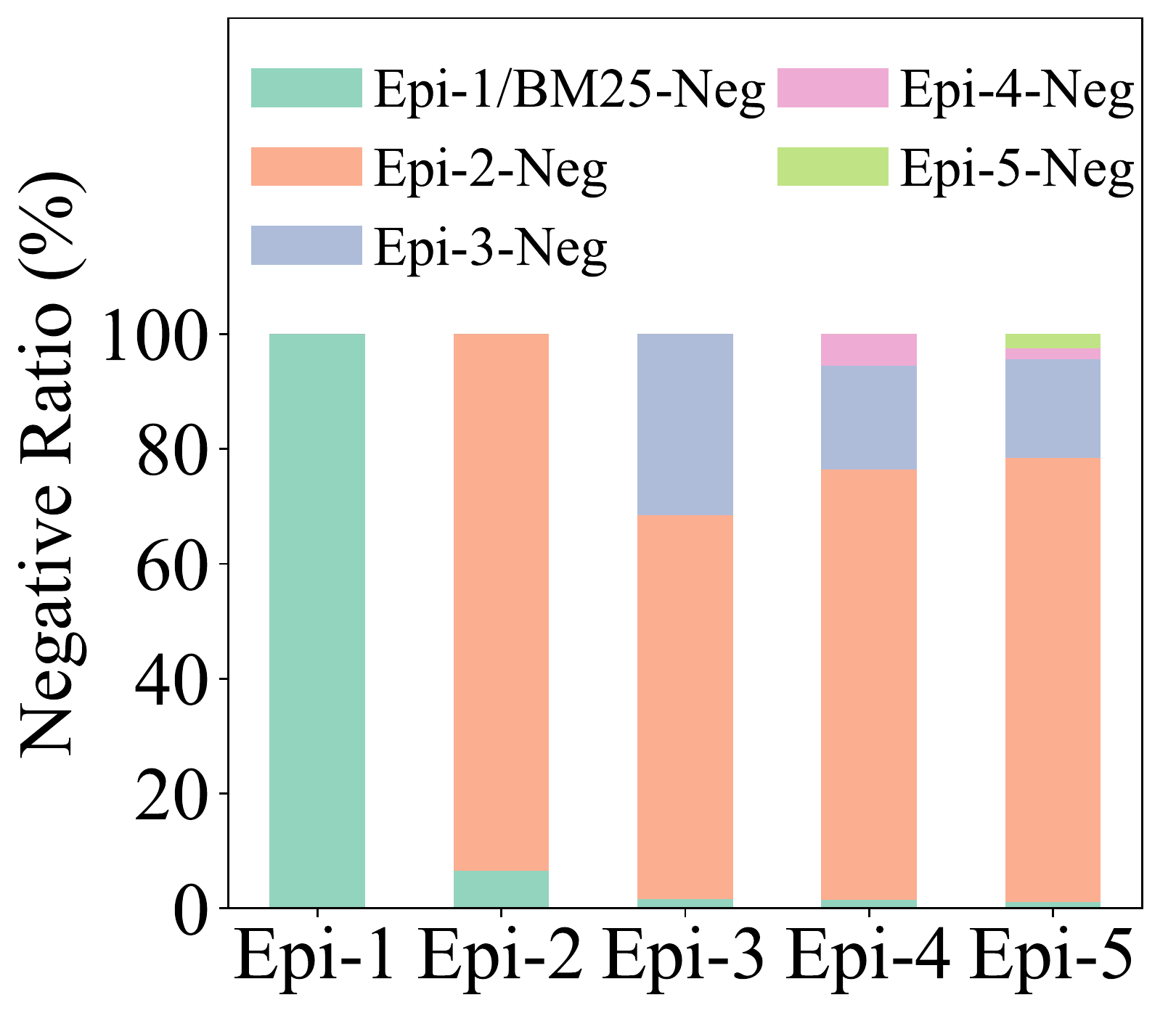}
        \centering
        \caption{Standard/coCondenser.\label{app-fig:stand-ance-coco}}
    \end{subfigure}
    \begin{subfigure}[t]{0.24\linewidth}
        \includegraphics[height=3.2cm]{Figures/app-ance-neg-dist/warmup-ance-bert-neg-ratio.pdf}
        \centering
        \caption{CyclicLR/BERT.\label{app-fig:ance-bert}}
    \end{subfigure} 
    \begin{subfigure}[t]{0.24\linewidth}
        \includegraphics[height=3.2cm]{Figures/ance-neg-dist/fig-ance-neg-ratio.pdf}
        \centering
        \caption{CyclicLR/coCondenser.\label{app-fig:ance-coco}}
    \end{subfigure} 
    \\
    \begin{subfigure}[t]{0.24\linewidth}
        \includegraphics[height=3.2cm]{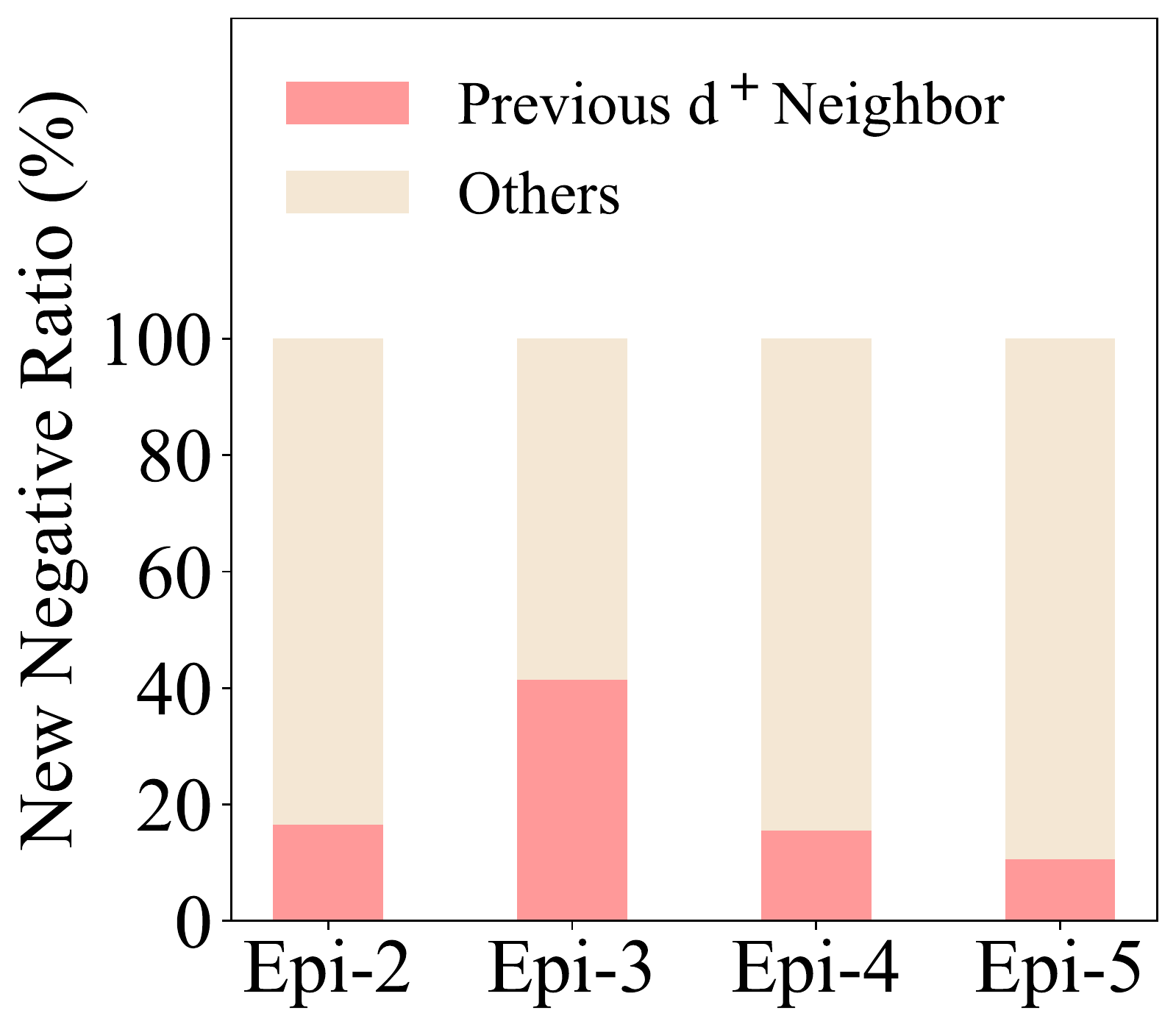}
        \caption{Standard/BERT.\label{app-fig:pos-stand-ance-bert}}
    \end{subfigure}
    \begin{subfigure}[t]{0.24\linewidth}
        \includegraphics[height=3.2cm]{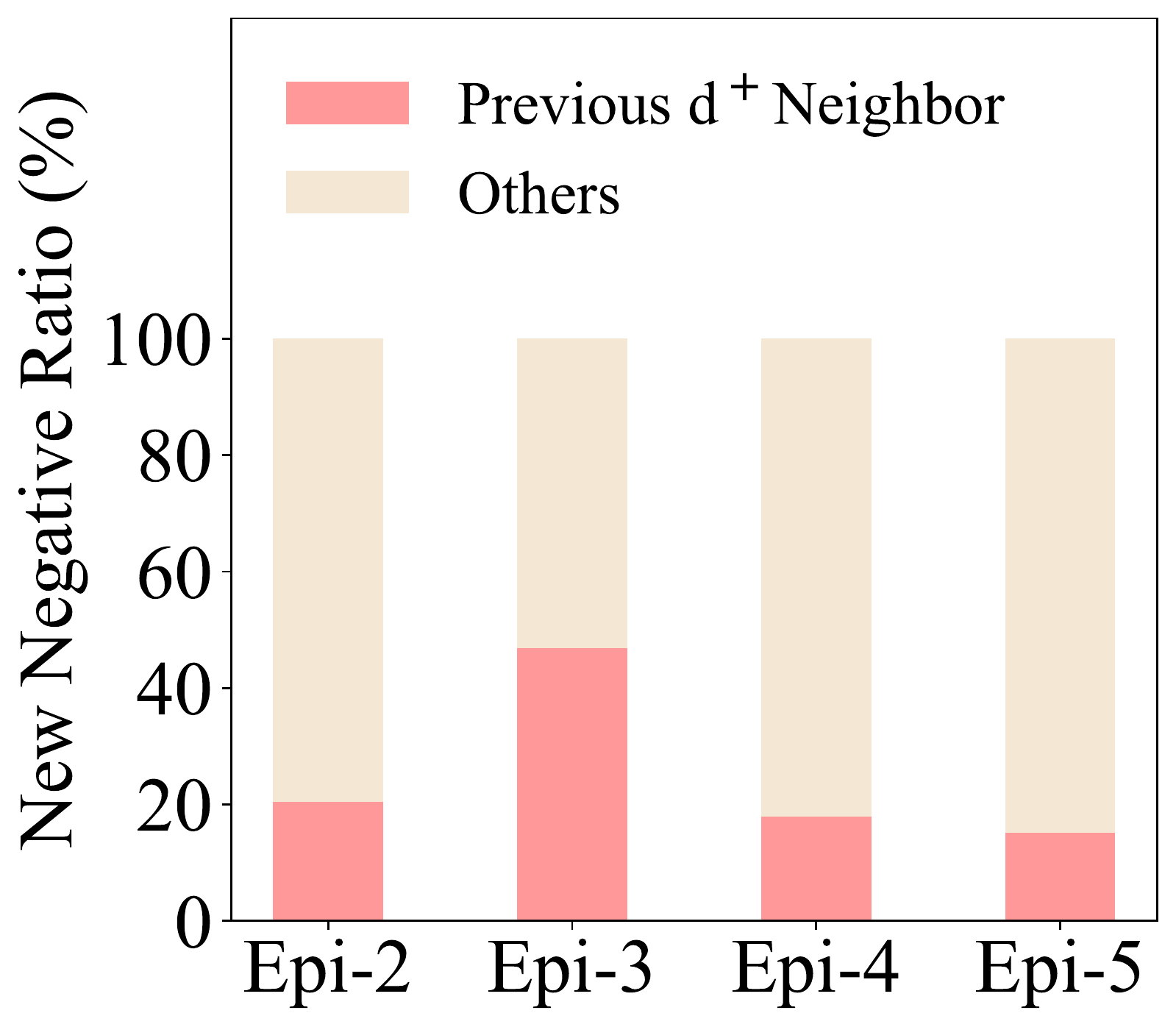}
        \caption{Standard/coCondenser.\label{app-fig:pos-stand-ance-coco}}
    \end{subfigure}
    \begin{subfigure}[t]{0.24\linewidth}
        \includegraphics[height=3.2cm]{Figures/app-ance-neg-dist/warmup-ance-bert-pos-ratio.pdf}
        \caption{CyclicLR/BERT.\label{app-fig:pos-ance-bert}}
    \end{subfigure}
    \begin{subfigure}[t]{0.24\linewidth}
        \includegraphics[height=3.2cm]{Figures/ance-neg-dist/fig-ance-pos-ratio.pdf}
        \caption{CyclicLR/coCondenser.\label{app-fig:pos-ance-coco}}
    \end{subfigure}
\caption{\label{app-fig:ance-neg-dist}Composition of training negatives for four ANCE variants. Figures (a) and (b) show the components of all training negatives for standard ANCE (BERT/coCondenser), and Figures (c) and (d) show the components of all training negatives for  ANCE (BERT/coCondenser) trained with CyclicLR. We exhibit the new negative constituent of standard ANCE (BERT/coCondenser) in Figures (e) and (f), and show the composition of new negatives for ANCE (BERT/coCondenser) trained with CyclicLR in Figures (g) and (h).
}
\end{figure*}

\section{Implementation Details}
\label{app:exp}

This section exhibits the detailed hyperparameters of ANCE-Tele and analyzes its training efficiency.

\subsection{Hyperparameters}
Table~\ref{app-tab:hyperpara} lists the detailed hyperparameters used by \model{} on the three evaluation datasets. The entire training process of \model{} consists of three episodes, labeled Epi-1, Epi-2, and Epi-3. Each episode trains the model from scratch based on the same initial model, using the same hyperparameters but different training negatives.
Take Epi-3 as an example: On MARCO, each training batch contains eight queries, each query is equipped with one positive and 31 negatives, and the number of training epochs is three. On NQ and TriviaQA, the query batch size is 128, each query is equipped with one positive and eleven negatives, and the total training epoch is 40. Next, we introduce how to obtain the training negatives for each episode. 


\subsection{Negative Mining}
The training negatives of \model{} come from newly-mined negatives and momentum negatives, where the newly-mined negatives include standard ANCE negatives and lookahead negatives. We set the ratio of newly-mined negatives to momentum negatives to 1:1 ($\alpha$=0.5). Specifically, Epi-1 uses the initial pre-trained model to mine new negatives, acting as the first training episode, so there is no negative momentum. To gather ANCE negatives and lookahead negatives for Epi-1, we first build two indexes for the query and positive passage separately, and then sample negatives from the top retrieved passages, where we keep the proportion of the two class negatives at 1:1 ($\beta$=0.5).

During iterative training, we employ an earlier negative-refreshing step because we find that refreshing negatives at an early stage achieves comparable performance and saves a lot of training time. For example, Epi-2 mines new negatives based on the early training checkpoints of Epi-1, where we set the negative-refreshing step as one-tenth of the total training steps without much tuning, i.e., 20k for MARCO and 2k for NQ and TrviaQA. In addition, Epi-2's momentum negatives are derived from the training negatives of the last episode (Epi-1).

\subsection{Training Efficiency}

Table~\ref{app-tab:train-time} shows the time cost of training \model{} in three episodes. Our implementation is initially based on Tevatron~\cite{tevatron} with modifications and has been integrated into OpenMatch~\cite{openmatch}.




\begin{table}[t]
\centering
\small
\resizebox{\columnwidth}{!}{
\begin{tabular}{l c c}
\toprule
\textbf{Methods} & \textbf{MRR@10}  & \textbf{R@1k} \\
\hline
ANCE-Tele & 39.1 & 98.4 \\
ANCE-Tele (continue training) & 39.2 & 98.3 \\
ANCE-Tele (w/ BM25 negatives) & 39.0 & 98.4  \\
ANCE-Tele (w/o overlap negatives) & 38.2 & 98.4 \\
\bottomrule
  \end{tabular}
}
\caption{The results of ANCE-Tele variants on MARCO. \label{app-tab:ance-tele-variants}} 
\end{table}

\section{Supplement to ANCE Results}
\label{app:ance-ana}

In this section, we first introduce the analysis setting of ACNE used in Section~\ref{sec:stable-analyses} and then investigate the negative composition for more ANCE variants.


\textbf{ANCE Steup.} The hyperparameters of ANCE remain the same as in the previous research~\cite{ance}. Table~\ref{app-tab:ance-param} shows the configurations. The warm-up method during iterative training is the difference between standard ANCE and ANCE with CyclicLR. Standard ANCE uses a single warm-up learning rate, while ANCE with CyclicLR uses a recurrent warm-up learning rate and achieves higher stability performance but slower convergence.



\begin{table}[t]
\centering
\resizebox{\columnwidth}{!}{
\begin{tabular}{l l c}
\toprule
\textbf{Episode} &
\textbf{Methods} & \textbf{Forget Rate (\%)} \\
\hline
\multirow{2}{*}{Epi-1} & ANCE-Tele & 8.9 \\
& ANCE-Tele (continue training) & 8.9 \\
\hline
\multirow{2}{*}{Epi-2} & ANCE-Tele & 18.5 \\
& ANCE-Tele (continue training) & 13.5 \\
\hline
\multirow{2}{*}{Epi-3} & ANCE-Tele & 15.9 \\
& ANCE-Tele (continue training) & 13.2 \\
\bottomrule
  \end{tabular}
}
  \caption{\label{app-tab:variant-forget} Forgetting rate of ANCE-Tele in different training modes.} 
\end{table}


\begin{table*}[ht]
\centering
\small
\begin{tabularx}{\textwidth}{p{3cm}|X|X}
\toprule
\textbf{Queries} & \textbf{Class A Negatives} & \textbf{Class B Negatives} \\ 
\hline
\textbf{\textit{(a)}} \redbf{most popular breed} of \bluebf{rabbit}
& 
The Golden Retriever is one of the \redbf{most popular breeds} in the United States. Learn more about this loveable dog with Golden Retriever facts \& pictures on petMD. 
& 
\bluebf{Rabbit} habitats include meadows, woods, forests, grasslands, deserts and wetlands. \bluebf{Rabbits} live in groups, and the best known species, the European \bluebf{rabbit}, lives in underground burrows, or \bluebf{rabbit} holes.
\\ \hline
\textbf{\textit{(b)}} what is the main difference between a \redbf{lightning strike} and \bluebf{static electricity}
& 
\redbf{Lightning} is a bright flash of electricity produced by a thunderstorm. All thunderstorms produce \redbf{lightning} and are very dangerous. If you hear the sound of thunder, then you are in danger from \redbf{lightning}. 
& 
\bluebf{Static Electricity}: Introducing Atoms This lesson lays the groundwork for further study of \bluebf{static} and current \bluebf{electricity} by focusing on the idea of positive and negative charges at the atomic level. 
\\ \hline
\textbf{\textit{(c)}} what \redbf{essential oil} to \bluebf{stop the itching of bug bites}
& 
\redbf{Essential Oils} to Treat Poison Ivy Rashes Having strong antiseptic properties, cypress \redbf{essential oil} is also among the best \redbf{essential oils} to treat poison ivy rashes.
&
Home Remedy to \bluebf{Stop Itching From Bug Bites} Home remedies treating \bluebf{itching from bug bites} vary in effectiveness, depending upon the type of \bluebf{bug bite} and the severity of the \bluebf{itching}.
\\ \hline
\textbf{\textit{(d)}} \redbf{how long} did the \bluebf{han dynasty} \redbf{last}
& 
\redbf{How long} did the Roman Empire \redbf{last}? From the time of Augustus to the time of Constantine Dragases, it was 1484 years.
& 
\bluebf{Han Dynasty} Achievements It was during the \bluebf{Han} period that contact with the West through the Silk Road was first established. 
\\ \hline


\textbf{\textit{(e)}} where are the \bluebf{pectoral} \redbf{muscles located}
& 
Biceps femoris Biceps femoris. The biceps femoris is a double-headed \redbf{muscle located} on the back of thigh. It consists of two parts: the long head, attached to the ischium (the lower and back part of the hip bone), and the short head, which is attached to the femur bone.
& 
\bluebf{Pectoralis} Major Muscle Function. The \bluebf{pectoralis} major muscle is the most important muscle for the adduction and anteversion of the shoulder joint which is why it is also known as the breaststroke muscle.
\\ \hline
\textbf{\textit{(f)}} how many \redbf{calories and carbs} in \bluebf{cantaloupe}
& 
Thomas' everything bagel - 280 \redbf{calories}, 3g of fat, and 53g of \redbf{carbs} per bagel. Visit our site for complete nutrition facts information for this item and 100,000+ additional foods. 
& 
10 Amazing Nutritional Benefits of \bluebf{Cantaloupe} 10 Amazing Health \& Nutritional Benefits of \bluebf{Cantaloupe}, Nutrition Facts of \bluebf{Cantaloupe}. \bluebf{Cantaloupe} is a delicious fruit with a unique flavor.
\\ \hline
\textbf{\textit{(g)}} who \redbf{invented} the \bluebf{shopping cart}
& 
The zoetrope was \redbf{invented} in 1834 by William Horner who called it a daedalum or daedatelum. However it is believed that Horner may have based his \redbf{invention} on that of a basic zoetrope created by a Chinese inventor.
& 
Your \bluebf{shopping cart} is empty! Founded in 1921 as Verlag Chemie, we can look back over 90 years of publishing in the fields of chemistry, material science, physics and life sciences as well as business and trade.
\\ \hline
\textbf{\textit{(h)}} when was \bluebf{houston} art \redbf{institute} founded 
& 
Cato \redbf{Institute} The Cato \redbf{Institute's} articles of incorporation were filed in December 1974, with the name Charles Koch Foundation, listing the original directors as Charles Koch, George Pearson, and Roger MacBride.
& 
\bluebf{Houston} City Council The \bluebf{Houston} City Council is a city council for the city of \bluebf{Houston} in the U.S. state of Texas. Currently, there are sixteen members, 11 elected from council districts and five at-large. 
\\ 
\bottomrule
\end{tabularx}
\caption{Cases of ANCE’s negative swing behavior in the iterative training process. We sample 8 MARCO training queries whose negatives show two classes of features during training: \redbf{Class A negative} represents the negative that comes in Epi-2 but goes out in Epi-3 and again in Epi-4. Conversely, \bluebf{Class B negative} comes in Epi-3 but goes out in Epi-4 and again in Epi-5. It is worth noting that these two types of negatives cover different aspects of the query.\label{app-tab:ance-swing-cases}
}
\end{table*}



\textbf{ANCE Variants.} Figure~\ref{app-fig:ance-neg-dist} presents the training negative composition of four ANCE variants, i.e., standard ANCE (BERT/coCondenser), ANCE trained with CyclicLR (BERT/coCondenser). Figure~\ref{app-fig:stand-ance-bert} to Figure~\ref{app-fig:ance-coco} show the component of the total training negatives, and Figure~\ref{app-fig:pos-stand-ance-bert} to Figure~\ref{app-fig:pos-ance-coco} show the composition of their new negatives.


Notably, BM25 negatives are quickly discarded for all variants after the first training episode, and the negative swing phenomenon is most evident in standard ANCE (BERT), as shown in Figure~\ref{app-fig:stand-ance-bert}. Moreover, we observe the phenomenon on all four variants, i.e., considerable new negatives of the current episode reside close to $d^+$ of the last episode.


\section{Supplement to ANCE-Tele Results}
\label{app:tele-variants}




Additionally, we provide the results of other ANCE-Tele variants using different training modes and negatives. Experiments are performed on MARCO, and the pre-trained model uses coCondenser, and the maximum training episode is set to three. The results are presented in Table~\ref{app-tab:ance-tele-variants}.

\textbf{ANCE-Tele (continue training).} ANCE adopts the \textit{continue training} mode --- the current episode takes the output model of the last episode as the initial model. By contrast, ANCE-Tele employs the \textit{training from scratch} mode, whereby each training episode initializes with the initial pre-train model coCondenser. For comparison, we also implement ANCE-Tele using the \textit{continue training} mode. As shown in Table~\ref{app-tab:ance-tele-variants}, the change in training mode makes little difference in performance, and even a slight MRR@10 improvement has been observed in the \textit{continue training} variant. Furthermore, we compare the forgetting rate between the two training modes and exhibit the results in Table~\ref{app-tab:variant-forget}. The results indicates that the \textit{continue training} mode further reduces the forgetting rate of ANCE-Tele.

\textbf{ANCE-Tele (w/ BM25 negatives).} We also test the effect of adding BM25 negatives to ANCE-Tele negatives. As expected, the addition of BM25 negatives brings no additional benefits, which shows little complementarity between ANCE-Tele negatives and BM25 negatives. 

\textbf{ANCE-Tele (w/o overlap negatives).} ANCE-Tele negatives from different mining sources (momentum/lookahead) share some overlaps. We thus test the effect of performing deduplication based on ANCE-Tele negatives. As shown in Table~\ref{app-tab:ance-tele-variants}, it is expected that negative deduplication significantly impacts ANCE-Tele. There are overlapping negatives because different mining sources refer to certain negatives multiple times, which can be considered significant negatives. Intuitively, keeping the overlapping part in the negative pool is equivalent to weighting them during training, which benefits the final performance.



\section{Case Studies}
\label{app:case}

Table~\ref{app-tab:ance-swing-cases} lists ten training cases on MARCO to visually show the negative swing behavior of ANCE. 

We observe two typical oscillating negative classes near the training query during iterative training, i.e., \redbf{Class A} and \bluebf{Class B}. Class A negatives first appear in Epi-2 (Top200 KNN), are discarded at Epi-3, and reappear in Epi-4 training negatives. On the contrary, Class B negatives first appear in Epi-3, are discarded at Epi-4 and reappear at Epi-5. The swing behavior of these two types of negative groups is precisely the opposite. More Interestingly, we observe that the two classes of hard negatives cover different aspects of the query.

For the query ``most popular breed of rabbit'' in case (a), the class A negative captures the semantics of ``most popular breed'' but the breed is ``dog'' rather than ``rabbit''. Instead, the class B negative alone captures  ``rabbit''. In case (b), the query asks the difference between lightning and static electricity; its class A negative alone captures ``lightning'', while its class B negative captures only ``static electricity''. In case (c), the query asks ``what essential oils can stop itching from bug bites'', the class A negative passage mentions ``essential oils'', but the passage is about treating poison ivy rashes. In contrast, the class B negative mentions the ``stop itching bug bites'' part but the key query term ``essential oils'' does not appear. Likewise, case (d) asks ``how long the Han dynasty lasted''. Its class A negative captures the duration of the dynasty but focuses on the Roman Empire; class B negative captures the ``Han dynasty'', but it describes the achievements of the Han dynasty rather than its duration. This phenomenon also exists in the remaining examples.


To sum up, these examples reveal that the model swings between the distinct classes of negatives during iterative training. In contrast, ANCE-Tele carries teleportation negatives to mitigate negative swings for better stability and convergence speed.



\section{Contributions}

Si Sun and Chenyan Xiong designed the methods and the experiments. Si Sun conducted the experiments. Chenyan Xiong and Si Sun wrote the paper. Yue Yu and Arnold Overwijk engaged in the discussion, revision, and response to reviewers. Zhiyuan Liu gave suggestions and feedback about the project and provided the experimental hardware. Jie Bao proofread the paper. 
\label{sec:appendix}


\end{document}